\documentclass[inkheadings]{phai}
\setcitestyle{numbers,square,comma,sort&compress}
\usepackage{tikz}
\usetikzlibrary{positioning,arrows.meta,shapes.geometric,fit,backgrounds,calc,decorations.pathreplacing}
\usepackage{mathtools}
\usepackage{algorithm}
\usepackage{algorithmic}
\usepackage{makecell}
\usepackage{tabularx}
\usepackage{enumitem}
\usepackage{pifont}
\usepackage{xurl}
\usepackage{verbatim}

\usepackage{float}
\usepackage{placeins}
\title{Discovery Foundation Models: \\Toward Open-Ended Discovery Intelligence}

\authors{%
  Ling Yang\quad
  Zhenfei Yin\quad
  Yingcheng Wu%
}

\runningtitle{Discovery Foundation Models}

\begin{document}
\maketitle

\vspace*{-2.2em}
{\centering
{\large\textbf{DFM Scientist Collaboration Program}}\\[4pt]
\begin{minipage}{0.92\linewidth}\centering\small
We work with scientists and experimental platforms on open problems with real
scientific value and real validation conditions. If you have such a problem,
or the data, code, compute, or lab conditions to investigate one, we would
like to hear from you.
\end{minipage}\\[8pt]
\setlength{\fboxsep}{5pt}%
\colorbox{black!5}{\href{https://phai-labs.com/collaborate/}{\large phai-labs.com/collaborate}}\\[8pt]
\begin{tabular}{@{}r@{\hspace{0.6em}}l@{}}
\small\textbf{Build with us} & \small\href{https://github.com/Gen-Verse/DFM-Plans}{github.com/Gen-Verse/DFM-Plans}\\
\small\textbf{Contact}       & \small\href{mailto:yang@phai-labs.com}{yang@phai-labs.com}\\
\end{tabular}\\[5pt]
\rule{0.75\linewidth}{0.4pt}\par}

\vspace{0.5em}

\begin{center}
  \includegraphics[width=0.99\linewidth]{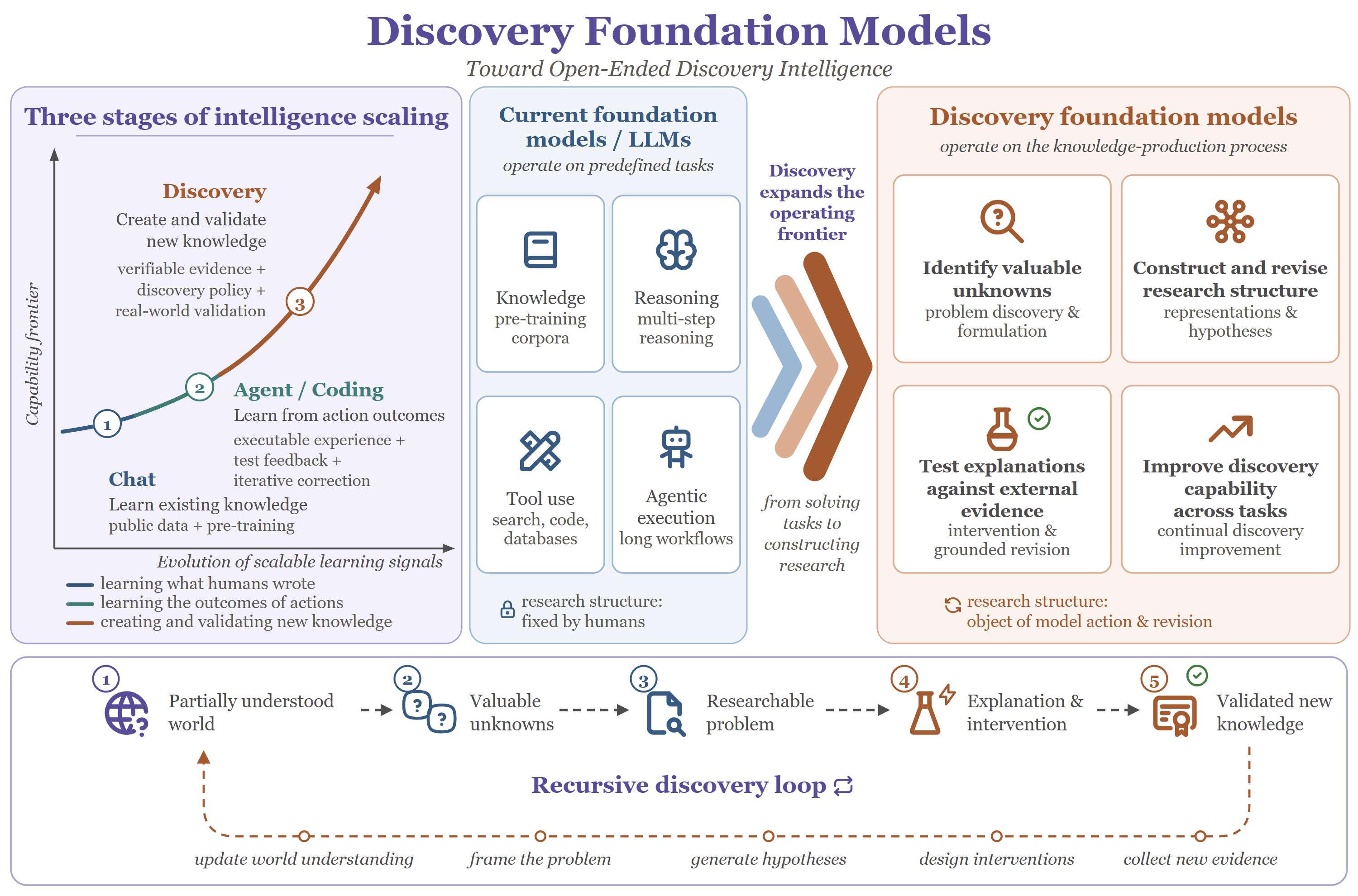}
  \captionof{figure}{\textbf{From task solving to discovery intelligence.}
The left panel illustrates three stages of intelligence scaling, from learning existing knowledge (\emph{Chat}), to learning from action outcomes (\emph{Agent/Coding}), and ultimately to creating and validating new knowledge (\emph{Discovery}). Current foundation models primarily operate over predefined tasks through knowledge, reasoning, tool use, and agentic execution, whereas \emph{Discovery Foundation Models} extend the operating frontier to the knowledge-production process itself: identifying valuable unknowns, constructing and revising research structure, testing explanations against external evidence, and improving discovery capabilities across tasks. The bottom panel depicts the resulting \emph{recursive discovery loop}, in which validated findings continually update world understanding and seed subsequent rounds of problem discovery, hypothesis generation, intervention, and evidence collection.}
  \label{fig:dfm-overview}
\end{center}

\newpage
\begin{abstract}
Foundation models have progressed from learning and reasoning over existing knowledge, to increasingly learning through action, tool use, and outcome feedback. We argue that the next frontier is a further transition: from solving and acting within problems specified by humans to participating in the process by which new problems, representations, explanations, and knowledge are created. We refer to this capability as \textbf{Discovery Intelligence}.
We formulate \textbf{Discovery Foundation Models (DFMs)} as general-purpose model systems for open-ended discovery. A DFM operates over a revisable research state and supports seven coupled capabilities spanning problem discovery, formulation, representation construction, hypothesis formation, intervention, evidence-grounded revision, and continual discovery improvement.
We instantiate this framework with \textbf{Zetema}, which couples explicit research-state dynamics, verification and experimental gating, external grounding, and cross-task Discovery Skill evolution. We further ground the framework with \textbf{GALILEO}, a real therapeutic-discovery system in which Dry-Lab reasoning, robotic and hands-on Wet-Lab experimentation, external biological evidence, and iterative hypothesis and design revision form a closed physical discovery loop. We then formulate a unified approach to capability formation and process-centered evaluation, enabling discovery behavior to be trained, improved, and measured beyond final-answer performance.
Together, these components establish discovery as a learnable, executable, and evaluable capability of foundation-model systems. We view this shift as a broader progression in intelligence scaling: from learning over existing knowledge, to learning from action outcomes, and ultimately to participating in the construction, testing, and revision of the structures through which new knowledge is discovered.
\end{abstract}



    

{\setcounter{tocdepth}{2}\tableofcontents}
\clearpage


\section{Introduction}
\label{sec:introduction}

Foundation models have become general interfaces to knowledge work. Large-scale pretraining, post-training, multimodal learning, coding, tool use, and agentic execution allow them to synthesize literature, reason over technical problems, analyze data, run software, and coordinate long workflows \citep{bommasani2021foundation,brown2020fewshot,wei2022cot,yao2023react,schick2023toolformer}. In science, these capabilities already support protein and materials modeling, weather prediction, mathematical and programmatic reasoning, literature-grounded analysis, and increasingly automated experimentation \citep{jumper2021alphafold,abramson2024alphafold3,merchant2023gnome,zeni2025mattergen,lam2023graphcast,price2025gencast,bran2024chemcrow,boiko2023coscientist,szymanski2023alab}.

Most of these systems begin from a research structure that people have already chosen. The question is stated, the variables are supplied, the objective is fixed, tools are exposed through an interface, and an evaluator determines whether the output is acceptable. Models can search and optimize inside this structure with increasing sophistication. They are much weaker when progress requires changing the structure itself.

That distinction matters in open-ended discovery. An apparent anomaly may be a measurement artifact. Two explanations can fit all existing observations because the available observable is non-identifying. A benchmark may reward a proxy. A persistent failure can result from a missing variable rather than a weak optimizer. In such cases, the next useful action is not another answer inside the current task. The system must decide what is actually unknown, how the problem should be posed, which representation makes competing mechanisms expressible, and what intervention could force them to disagree \citep{scholkopf2021causalrep,brunton2016sindy,udrescu2020aifeynman,chaloner1995bayesdesign}.

We refer to this broader target as \emph{Discovery Intelligence}. Generalist problem solving asks how broadly and deeply a model can solve supplied tasks. Discovery Intelligence asks whether a model system can construct, test, and revise the process through which a partially understood world becomes validated knowledge. The distinction is increasingly visible in scientific-agent evaluations: long research workflows can be executed successfully while evidence integration, refutation-driven revision, and long-horizon reliability remain fragile \citep{riosgarcia2026scientists,garikaparthi2026researchgym}.

Science is a useful capability-forming environment for this target because it exposes incomplete specifications that ordinary benchmarks often remove. Questions can be underspecified, variables hidden, mechanisms observationally equivalent, interventions costly, evaluators incomplete, and outcomes delayed. Evidence also arrives from environments that the model cannot rewrite after seeing the result. These properties turn formulation, representation, experiment design, failure attribution, and revision into observable decisions rather than rhetorical qualities of a final answer \citep{wang2023scientificdiscovery,zhang2025scientificmethod,swanson2025virtuallab,gottweis2026coscientist,ghareeb2026robin,lu2026aiscientist,trost2026spark}.

We introduce \emph{Discovery Foundation Models} as a model-system category for this setting. A DFM identifies valuable unknowns, formulates researchable problems, constructs and revises representations, forms testable explanations, designs informative interventions, updates the research state from external evidence, and improves these operations across tasks and domains. The category is broader than hypothesis generation and different from simply applying a foundation model to scientific data. It concerns which parts of knowledge production are fixed inputs and which can become objects of model action and revision \citep{wang2024scimon,baek2025researchagent,gottweis2026coscientist}.

We then instantiate the framework with \emph{Zetema}. Zetema maintains an explicit research state, supports branching and rollback within an investigation, gates consequential actions through verification and a Research World Model, connects Dry-Lab reasoning to external computational or physical evidence, and converts validated cross-task experience into Discovery Skills. This organization makes the proposed capability operational without requiring one monolithic model or maximal autonomy.

We additionally connect the framework to a real Dry-Lab/Wet-Lab discovery case. GALILEO couples multi-omics-informed target nomination and peptide design with robotic synthesis, multimodal phenotyping, orthogonal hands-on assays, and repeated evidence-driven revision. Across experimentally validated LRRC8C and SLC25A1 branches, physical measurements alter subsequent target beliefs, assay choices, mechanism hypotheses, and molecular-design policies; across five optimization rounds, the resulting feedback is further consolidated into a transferable Amphiphilic Balance Grammar. We use this case as empirical grounding for the intervention--evidence--revision loop, while keeping the broader general-purpose DFM claim distinct from any single domain-specific system.

The learning and evaluation formulations follow the same state-centered view. Training targets the intermediate decisions that change a research program, using trajectories, interactive environments, process supervision, scientific feedback, and resource allocation across formulation, representation, hypothesis construction, intervention, falsification, and verification. Evaluation measures both externally validated progress in the current episode and improvement in future discovery behavior under matched resources and retrieval controls \citep{majumder2025discoverybench,chen2025scienceagentbench,huang2024mlagentbench,chan2025mlebench,starace2025paperbench,song2025sde}.

\paragraph{Contributions.}
This paper makes three contributions.
\begin{itemize}[leftmargin=1.4em, itemsep=2pt, topsep=3pt]
    \item We formulate Discovery Foundation Models as a capability-based model-system category and specify the research objects and operations that distinguish open-ended discovery from optimization over a predefined task.
    \item We define the Discovery Process and instantiate it with Zetema, which couples explicit research-state revision, evidence-based action gating, external grounding, and validated cross-task Discovery Skill evolution.
    \item We formulate training and evaluation mechanisms for learning these discovery operations, allocating resources across the process, and measuring externally validated knowledge progress and transferable improvement on unseen tasks.
    \item We empirically ground the Dry-Lab/Wet-Lab component with GALILEO, a real therapeutic-discovery loop in which physical biological feedback revises subsequent scientific decisions and is distilled across rounds into a reusable design rule.
\end{itemize}

Sections~\ref{sec:motivation}--\ref{sec:discovery-process} introduce the problem setting and discovery operators. Sections~\ref{sec:zetema}--\ref{sec:evaluation} specify the system instantiation, capability formation, and evaluation protocol. Sections~\ref{sec:research-horizons} and~\ref{sec:governance} analyze how grounding and responsibility change as the same framework moves from digital to physical and recursive settings.

\section{From Generalist Problem Solving to Discovery Intelligence}
\label{sec:motivation}

The motivation for DFMs is not that current foundation models lack scientific knowledge or reasoning. Their limitation is more specific: most training and evaluation pipelines reward competence after the research structure has been fixed. 

\subsection{Generalist Capability within Predefined Research Structures}
\label{subsec:generalist-intelligence}

Foundation models have expanded from language modeling to broad knowledge, multi-step reasoning, coding, multimodal interaction, tool use, and agentic execution \citep{brown2020fewshot,wei2022cot,yao2023react,schick2023toolformer,guo2025deepseek}. Scientific models extend the same substrate to proteins, molecules, materials, physical fields, biomedical records, and other domain-specific modalities \citep{jumper2021alphafold,abramson2024alphafold3,merchant2023gnome,zeni2025mattergen,lam2023graphcast,price2025gencast}. Scientific agents connect these abilities to search, code, databases, simulators, and laboratory interfaces \citep{bran2024chemcrow,boiko2023coscientist,swanson2025virtuallab,gottweis2026coscientist,ghareeb2026robin}.

This capability is a necessary substrate for discovery, but its usual task interface hides a structural ceiling. A model receives a recognizable object---a question, dataset, benchmark, formal language, design space, or goal---and optimizes within it. Search can explore enormous candidate spaces, reinforcement learning can discover unexpected strategies, and an agent can automate a long workflow. None of these mechanisms guarantees that the supplied variables or evaluator are scientifically adequate.

A fixed representation cannot express a variable it omits. A search objective cannot recover a property that its evaluator systematically ignores. Increasing sample count does not distinguish mechanisms when the observable is non-identifying. An automated workflow can therefore pursue a misframed question more efficiently without becoming better at recognizing the misframing.

The boundary is easiest to see when a research program stalls. If two mechanisms remain observationally equivalent, the bottleneck may be the measurement rather than hypothesis diversity. If performance gains disappear under another data split, the problem may be evaluator mismatch rather than optimization. If every explanation requires local exceptions in the same regime, another representation may be more useful than another candidate explanation. These are changes to the research structure, not additional solutions inside it.

\subsection{Science as a Capability-Forming Environment}
\label{subsec:science-as-catalyst}

Scientific discovery exposes these structural decisions because evidence is coupled to a world that pushes back. The system must often act before it knows the correct question, choose measurements under partial observability, and revise after outcomes that do not match its predictions. Active intervention changes what can be learned: a perturbation, counterexample, boundary test, simulation, or replication can separate explanations that observational data leave equivalent \citep{chaloner1995bayesdesign,boiko2023coscientist,szymanski2023alab}.

This feedback is qualitatively different from adding more scientific text to pretraining. Knowledge helps a model recognize established concepts and plausible mechanisms; a capability-forming environment requires it to make consequential research decisions under incomplete specification. The environment can be a codebase, formal system, causal simulator, digital twin, robotic platform, physical laboratory, or human-mediated process. What matters is that the resulting observation is not freely chosen by the model.

Science also makes evaluator incompleteness visible. Benchmark leakage, simulator artifacts, non-reproducible effects, and publication-like plausibility can all create apparent progress without stronger knowledge. Work on AI-assisted science has already highlighted the risk of fluent but weakly grounded understanding and the possibility that AI changes which problems are pursued, not only how quickly they are solved \citep{messeri2024illusions,hao2026impact}. For a DFM, the evaluator can itself become part of the research state when evidence suggests that it is misaligned.

Long horizons make the training signal harder but more informative. A negative result may eliminate months of future work. A failed replication can reduce confidence in the phenomenon rather than in a particular hypothesis. A representation change can make later interventions identifying. These outcomes cannot be valued reliably from the final answer alone; they require a record of how the research state changed.

\subsection{Three Missing Transitions}
\label{subsec:missing-discovery-capabilities}

The gap between predefined problem solving and Discovery Intelligence can be localized to three transitions.

\paragraph{Framing.}
The system must move from observations and uncertainty to a research opportunity worth pursuing. This includes distinguishing persistent structure from noise, deciding which unknowns are consequential and testable, and specifying the scope, scale, conditions, and observables needed to make the problem researchable. A supplied question can be rejected or reformulated when it is too broad, proxy-driven, or impossible to identify under the available measurements.

\paragraph{Modeling.}
The system must construct the variables and abstractions through which explanations become expressible. A useful operation may add a latent variable, remove a proxy, change scale, separate regimes, revise an ontology, or transform the problem into a causal, geometric, symbolic, or programmatic form \citep{brunton2016sindy,udrescu2020aifeynman,scholkopf2021causalrep}. Hypotheses are then formed inside this provisional representation and must differ in mechanism, validity conditions, or intervention response rather than only in wording.

\paragraph{Grounding and revision.}
The system must choose evidence that can change the status of the current explanations and then update the appropriate research object. A contradiction can indicate theory failure, measurement error, protocol deviation, hidden confounding, simulator misspecification, or environmental shift. Discovery therefore requires both informative intervention and failure attribution. The resulting experience becomes a transferable Discovery Skill only after its trigger and effect survive validation beyond the episode in which it was observed.

These transitions define the objects that Sections~\ref{sec:definition} and~\ref{sec:discovery-process} make explicit.

\section{Discovery Foundation Models: Defining Discovery Intelligence}
\label{sec:definition}

\begin{figure}[!htbp]
  \centering
  \includegraphics[width=\linewidth]{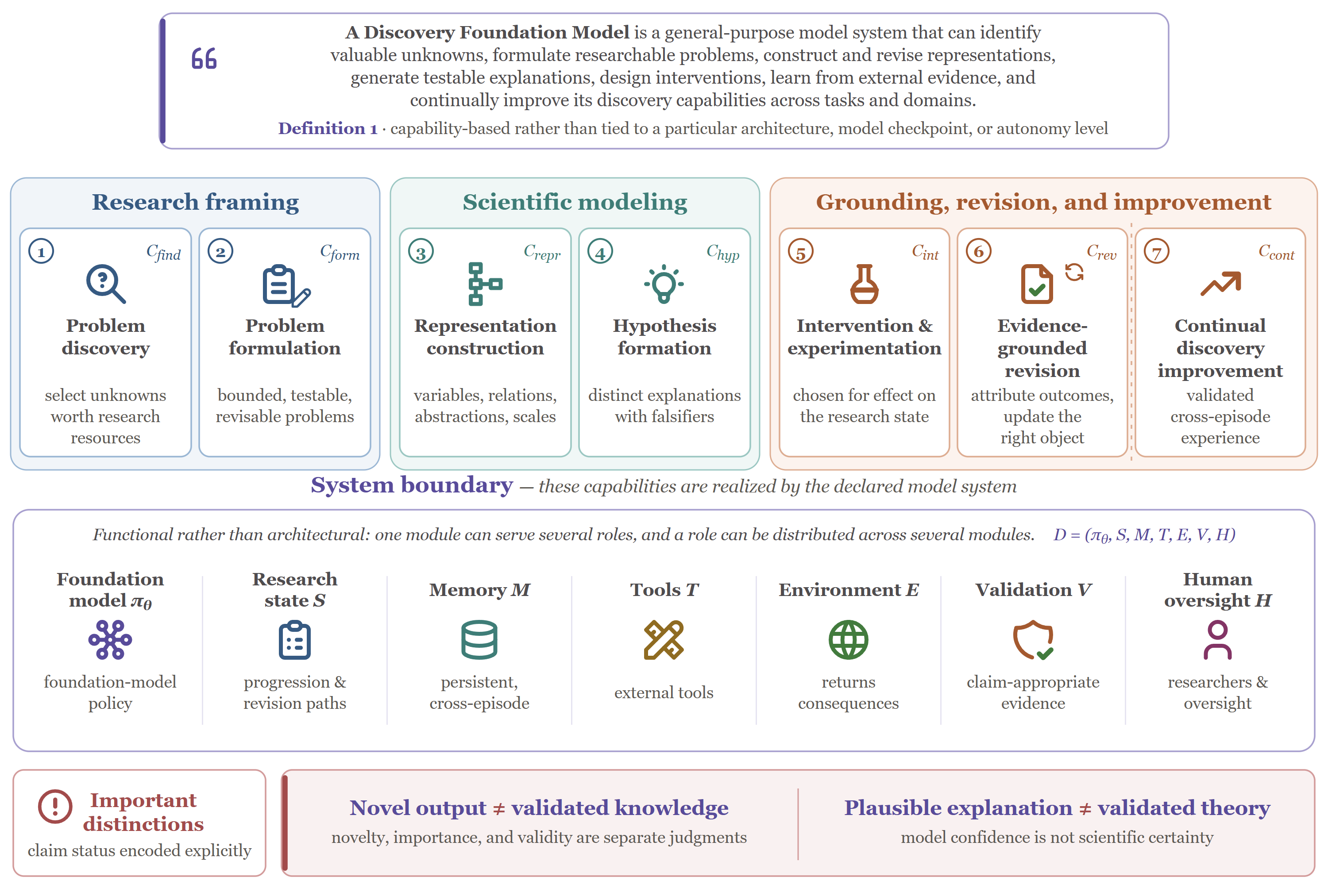}
\caption{
\textbf{Capability definition and system boundary of a Discovery Foundation Model.}
A DFM is defined by seven coupled capabilities spanning research framing, scientific modeling, evidence-grounded intervention and revision, and continual discovery improvement. These capabilities are realized by an integrated model system comprising a foundation-model policy, explicit research state, memory, tools, environments, validation mechanisms, and human oversight. The definition is therefore capability-based rather than tied to a particular architecture, model checkpoint, or autonomy level.
}
\label{fig:dfm-definition}
\end{figure}

We define DFMs by the research structures they can construct and revise, not by a particular neural architecture or degree of autonomy. Figure~\ref{fig:dfm-definition} summarizes the capability boundary.

\subsection{Problem Setting and Formal Definition}
\label{subsec:formal-definition}

A conventional model task can be abstracted as

\begin{equation}
\mathcal{Q}=(P,R,G,T,V),
\label{eq:predefined-task}
\end{equation}

where $P$ is the problem, $R$ its representation, $G$ the objective, $T$ the available tools, and $V$ the evaluator. The system is asked to produce a solution under this supplied structure. This abstraction covers scientific question answering, formal reasoning, tool-using agents, and search-based design even when the underlying task is difficult or the resulting solution is genuinely novel.

Discovery starts from a less complete state. The system observes a partially understood world $\mathcal{W}$, has an initial knowledge state $\mathcal{K}_0$, and operates under computational, experimental, safety, and access constraints $\mathcal{B}$. A discovery episode produces both knowledge progress and an evidence-bearing record of how the investigation changed:

\begin{equation}
\Phi_{\mathrm{disc}}:
(\mathcal{W},\mathcal{K}_0,\mathcal{B})
\longrightarrow
(\Delta\mathcal{K},\Xi).
\label{eq:discovery-episode}
\end{equation}

$\Delta\mathcal{K}$ denotes externally validated progress, while $\Xi$ contains the state transitions, alternatives, interventions, observations, failed formulations, and revisions that produced it. Unlike Equation~\ref{eq:predefined-task}, the problem, representation, hypothesis space, intervention strategy, and validation procedure can all change during $\Phi_{\mathrm{disc}}$.

\begin{quote}
\noindent
\textbf{Definition 1 (Discovery Foundation Model).} A \emph{Discovery Foundation Model} is a general-purpose model system that can identify valuable unknowns, formulate researchable problems, construct and revise representations, generate testable explanations, design interventions, learn from external evidence, and continually improve its discovery capabilities across tasks and domains.
\end{quote}

The definition imposes three requirements. First, the relevant operations must transfer beyond one fixed task even when their implementation remains domain-specific. Second, scientific claims are grounded by evidence appropriate to the domain; model confidence or internal agreement is not sufficient. Third, the evaluated unit is the declared model system, including any persistent state, memory, tools, environments, validators, and human participation that materially determine its behavior.

A DFM can therefore make useful progress without producing a final positive discovery. Showing that an effect does not replicate, that a question is untestable under current measurements, or that a representation omits the variable needed for intervention can all be valid outputs when the conclusion is supported by the research state.

\subsection{Discovery Capabilities}
\label{subsec:necessary-capabilities}

We factor the DFM target into seven coupled capabilities:

\begin{equation}
\mathcal{C}_{\mathrm{DFM}}
=
\{C_{\mathrm{find}},C_{\mathrm{form}},C_{\mathrm{repr}},C_{\mathrm{hyp}},C_{\mathrm{int}},C_{\mathrm{rev}},C_{\mathrm{cont}}\}.
\label{eq:dfm-capabilities}
\end{equation}

$C_{\mathrm{find}}$ selects unresolved structures worth allocating research resources to and rejects apparent unknowns that disappear under calibration, retrieval, or stronger baselines. $C_{\mathrm{form}}$ turns a selected unknown into a bounded and testable problem by fixing its object, scope, scale, conditions, and observables while keeping those choices revisable.

$C_{\mathrm{repr}}$ constructs the variables, relations, abstractions, and scales through which the problem is expressed. It becomes decisive when the current representation makes every candidate explanation equivalent or repeatedly produces the same failure boundary \citep{brunton2016sindy,udrescu2020aifeynman,scholkopf2021causalrep}. $C_{\mathrm{hyp}}$ forms mechanistically distinct explanations with explicit assumptions, validity ranges, predictions, and possible falsifiers \citep{wang2024scimon,baek2025researchagent,gottweis2026coscientist}.

$C_{\mathrm{int}}$ chooses experiments, simulations, code executions, ablations, counterexamples, alternative measurements, or replications for their expected effect on the research state rather than for confirmation alone. $C_{\mathrm{rev}}$ attributes unexpected outcomes and updates the appropriate object: hypothesis, representation, problem formulation, protocol, measurement process, or intervention plan.

$C_{\mathrm{cont}}$ changes future discovery behavior using validated cross-episode experience. This is stronger than fact accumulation or retrieving a successful trajectory. A reusable operation must specify when it applies, what it should change, and what later evidence would show that the change was beneficial. Memory-based agents provide precedents for experience-driven behavioral change; the DFM requirement adds attribution, scientific grounding, and transfer \citep{shinn2023reflexion,wang2024voyager}.

The first six capabilities operate within an investigation. The seventh is a cross-task update mechanism. Section~\ref{sec:discovery-process} specifies the within-task operators, and Section~\ref{sec:zetema} instantiates both levels in one system organization.

\subsection{System Boundary}
\label{subsec:model-systems}

A DFM is evaluated as a model system

\begin{equation}
\mathcal{D}=(\pi_{\theta},\mathcal{S},\mathcal{M},\mathcal{T},\mathcal{E},\mathcal{V},\mathcal{H}),
\label{eq:dfm-system}
\end{equation}

where $\pi_\theta$ is the foundation-model policy, $\mathcal{S}$ the research state, $\mathcal{M}$ persistent memory, $\mathcal{T}$ external tools, $\mathcal{E}$ the environment that returns consequences, $\mathcal{V}$ validation mechanisms, and $\mathcal{H}$ human researchers or oversight. The tuple is functional rather than architectural: one module can serve several roles, and a role can be distributed across several modules.

The system boundary matters for attribution. If a human supplies the decisive reformulation, the trajectory should record that intervention rather than attributing the discovery to the model. If one baseline receives a hand-built representation unavailable to another, the comparison is not a model-only comparison. Autonomy is similarly orthogonal to capability: a system can autonomously execute low-risk code while requiring approval for a physical experiment and still instantiate the same discovery operators.

Validation is not reduced to a universal scalar reward. Logical checks, held-out execution, simulation, replication, independent reviewers, and physical measurement support different claims. The system should preserve which validator supported which state transition and abstain or escalate when the available evidence does not justify promotion of a claim.

\subsection{Relation to Existing Scientific AI Paradigms}
\label{subsec:existing-paradigms}

DFMs build on, rather than replace, existing scientific AI. Foundation models for science provide domain representations and knowledge \citep{jumper2021alphafold,abramson2024alphafold3,merchant2023gnome,zeni2025mattergen}. Scientific reasoning models improve formal inference and verification \citep{wei2022cot,lightman2024verify,snell2025testtime,guo2025deepseek}. Scientific agents coordinate tools and long workflows \citep{baek2025researchagent,skarlinski2024paperqa2,gottweis2026coscientist,lu2026aiscientist,ghareeb2026robin}. Search systems explore candidate spaces, and autonomous laboratories connect proposals to physical measurements \citep{abolhasani2023selfdriving,boiko2023coscientist,szymanski2023alab,swanson2025virtuallab}.

The defining difference is the joint capability requirement. Let $\mathcal{Z}$ denote a predefined search space and $z\in\mathcal{Z}$ a candidate solution or design. Search can be central to a DFM, but optimizing candidates within a fixed $\mathcal{Z}$ under an evaluator $V$ does not by itself establish the ability to revise the research structure. A DFM must also be able to recognize, from evidence, when the search space or evaluator is inadequate and revise the relevant structure accordingly. Likewise, physical execution provides strong external grounding but does not by itself constitute discovery when the objective and design space remain human-specified. An AI-scientist system satisfies the DFM criterion to the extent that formulation, representation, intervention, and revision become explicit model-system operations, their consequences are externally grounded, and the resulting experience is validated to improve future discovery behavior across tasks.

\begin{table*}[t]
\centering
\caption{Capability-level comparison of Discovery Foundation Models and existing
scientific AI paradigms.}
\label{tab:dfm-comparison}

\footnotesize
\setlength{\tabcolsep}{3pt}
\renewcommand{\arraystretch}{1.25}

\begingroup
\hyphenpenalty=10000
\exhyphenpenalty=10000

\newcommand{\tablehead}[2]{%
  \begin{minipage}[c][1.05cm][c]{\linewidth}
    \centering\scriptsize
    \textbf{#1}\\
    \textbf{#2}
  \end{minipage}%
}

\begin{tabularx}{\textwidth}{
>{\raggedright\arraybackslash}p{2.7cm}
*{6}{>{\centering\arraybackslash}X}
}
\toprule

\begin{minipage}[c][1.05cm][c]{\linewidth}
  \raggedright
  \textbf{Paradigm}
\end{minipage}
&
\tablehead{Problem}{Source}
&
\tablehead{Representation}{Change}
&
\tablehead{Intervention}{Design}
&
\tablehead{Problem}{Revision}
&
\tablehead{External}{Grounding}
&
\tablehead{Skill}{Transfer}
\\

\midrule

\makecell[l]{Foundation Models\\for Science}
&
Given
&
Limited
&
External
&
Not defining
&
Data
&
Not defining
\\

\addlinespace[2pt]

\makecell[l]{Scientific Reasoning\\Models}
&
Given
&
Limited
&
Limited
&
Not defining
&
\makecell[c]{Tasks or\\verifiers}
&
Not defining
\\

\addlinespace[2pt]

\makecell[l]{Scientific Agents and\\AI Scientist Systems}
&
\makecell[c]{Often\\given}
&
Optional
&
Often
&
Optional
&
\makecell[c]{Tools and\\environments}
&
\makecell[c]{Usually\\task-specific}
\\

\addlinespace[2pt]

\makecell[l]{Search and\\Evaluator Systems}
&
Given
&
Fixed
&
\makecell[c]{Domain\\dependent}
&
Not defining
&
Evaluator
&
Not defining
\\

\addlinespace[2pt]

\makecell[l]{Autonomous\\Laboratories}
&
\makecell[c]{Often\\given}
&
\makecell[c]{Task\\defined}
&
Required
&
Optional
&
Physical
&
Not defining
\\

\midrule

\textbf{\makecell[l]{Discovery Foundation\\Models}}
&
\textbf{Constructed}
&
\textbf{Required}
&
\textbf{Required$^{*}$}
&
\textbf{Required}
&
\textbf{Required}
&
\textbf{Required}
\\

\bottomrule
\end{tabularx}

\endgroup

\vspace{2pt}

\begin{minipage}{\textwidth}
\footnotesize
\textit{Note.}
Entries describe the typical capability targets of each paradigm rather than
universal properties of every individual system. ``Not defining'' means that
the capability is not required by the paradigm. ``Optional'' means that some
systems support the capability without treating it as a defining requirement.
$^{*}$Intervention design is required when the research setting permits
experimental, computational, or other active forms of evidence acquisition.
\end{minipage}

\end{table*}

The \cref{tab:dfm-comparison} describes typical capability targets rather than mutually exclusive categories. A particular scientific agent may already reconstruct representations, and a search system may alter its evaluator. Such systems satisfy more of the DFM definition to the extent that these operations are integrated, externally grounded, and evaluated as transferable behavior. Conversely, the DFM label does not supply capabilities that the system has not demonstrated.

\section{Discovery Process: Operationalizing Discovery Intelligence}
\label{sec:discovery-process}

\begin{figure}[!htbp]
  \centering
  \includegraphics[width=\linewidth]{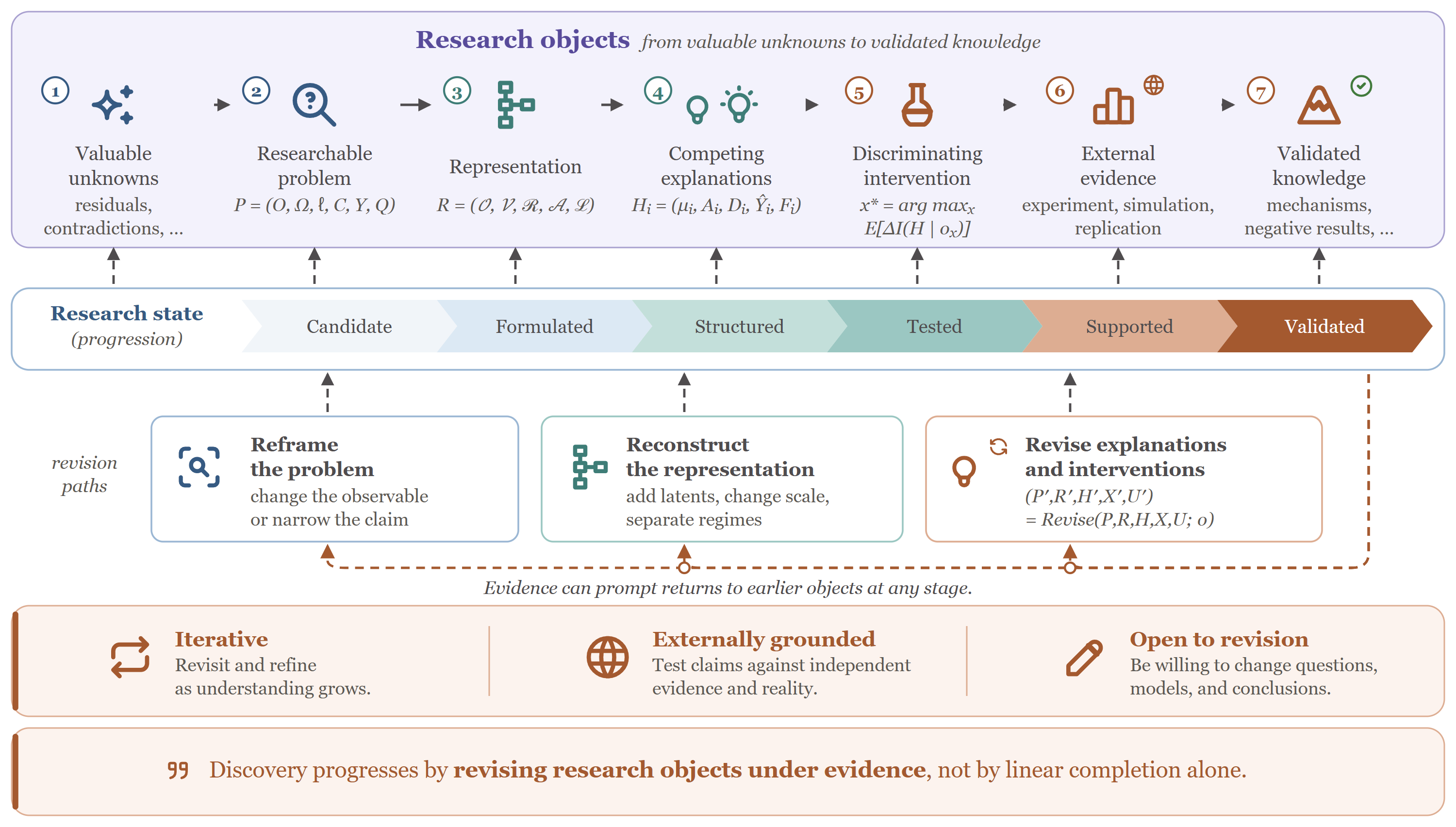}
\caption{
\textbf{The Discovery Process as revision over an evolving research state.}
Discovery progresses from valuable unknowns to researchable problems, representations, competing explanations, discriminating interventions, external evidence, and validated knowledge. These objects are not traversed as a fixed pipeline: evidence can return the investigation to an earlier problem, representation, explanation, or intervention. The research state therefore records both epistemic progression and the revision paths through which that progression is achieved.
}
\label{fig:discovery-process}
\end{figure}

The DFM definition specifies \emph{what} the system must be able to revise. The Discovery Process specifies \emph{how} those revisions compose inside an investigation. Figure~\ref{fig:discovery-process} shows the main research objects and feedback paths.

The process is not a fixed stage pipeline. Several formulations may coexist, different representations can support different hypothesis families, and evidence can return the investigation to an earlier object. The unit of computation is therefore a transition in a revisable research state rather than a completed textual stage.

\subsection{From Unknowns to Researchable Problems}
\label{subsec:valuable-unknowns}

Problem Discovery begins from observations, residuals, contradictions, failed replications, regime boundaries, evaluator mismatch, or newly available measurements. The operator filters as well as proposes. A signal that disappears after calibration, stronger retrieval, or a more appropriate baseline should not be promoted into a research program merely because it was initially surprising.

A selected unknown is then formulated as

\begin{equation}
P=(O,\Omega,\ell,C,Y,Q),
\label{eq:research-problem}
\end{equation}

where $O$ is the object of study, $\Omega$ its scope, $\ell$ the relevant scale, $C$ the conditions under which the claim is posed, $Y$ the observables, and $Q$ the unresolved relation or mechanism. These fields are operational: they determine which evidence can count, which interventions are feasible, and where the resulting knowledge is expected to apply.

Formulation becomes an active discovery step when alternatives imply different experiments. The same observation may be treated as a prediction failure, a causal-identification problem, or a measurement problem. If no available intervention can resolve the central uncertainty under one framing, the correct update may be to change the observable or narrow the claim rather than to continue searching for answers inside that framing.

\subsection{Representations and Competing Explanations}
\label{subsec:scientific-representations}

Given a provisional problem, the system constructs a scientific representation

\begin{equation}
R=(\mathcal{O},\mathcal{V},\mathcal{R},\mathcal{A},\mathcal{L}),
\label{eq:scientific-representation}
\end{equation}

with objects $\mathcal{O}$, variables $\mathcal{V}$, relations $\mathcal{R}$, abstractions or coarse-grainings $\mathcal{A}$, and structural constraints $\mathcal{L}$. Representation operations include adding a latent variable, removing a misleading proxy, changing temporal or spatial scale, separating regimes, revising an ontology, or translating the problem into another formal structure \citep{brunton2016sindy,udrescu2020aifeynman,scholkopf2021causalrep}.

A representation earns its role through downstream consequences. It should improve prediction under unseen conditions, expose a discriminating intervention, separate previously conflated regimes, compress a mechanism, or transfer to another setting. A novel label that leaves every possible action unchanged is not a useful representation change.

Within a provisional representation, a candidate explanation is

\begin{equation}
H_i=(\mu_i,A_i,D_i,\widehat{Y}_i,F_i),
\label{eq:scientific-hypothesis}
\end{equation}

where $\mu_i$ is the mechanism, $A_i$ its assumptions, $D_i$ its validity range, $\widehat{Y}_i$ its predictions, and $F_i$ the observations that would weaken or falsify it. Candidate explanations are useful when they disagree under at least one relevant condition. If every current hypothesis predicts the same observation under every feasible action, additional hypothesis sampling is unlikely to be the bottleneck; the system should inspect the representation or measurement interface.

Explanations can also be nested. A high-level regularity can remain valid after its proposed lower-level mechanism fails, and different mechanisms can dominate in different regimes. The research state therefore records which claim is under test rather than forcing all explanations into one winner-take-all competition.

\subsection{Intervention and Evidence-Grounded Revision}
\label{subsec:discriminating-interventions}

An intervention is selected for the state change it is expected to produce. Let $H$ denote the current explanation set and $o_x$ the possible observation under intervention $x$. A conceptual objective is

\begin{equation}
x^{*}
=
\operatorname*{arg\,max}_{x}
\mathbb{E}[\Delta\mathcal{I}(H\mid o_x)],
\label{eq:discriminating-intervention}
\end{equation}

where $\Delta\mathcal{I}$ measures the expected reduction or restructuring of uncertainty over the current explanations. In practice, the decision also depends on feasibility, cost, risk, statistical power, measurement quality, and the probability of an inconclusive outcome \citep{chaloner1995bayesdesign,gandhi2025boxinggym}.

The selected action can be a physical experiment, simulation, code execution, ablation, counterexample, alternative measurement, or replication. Confirmation is not the only useful outcome. An intervention can reveal that all current hypotheses share a false assumption, that the manipulation failed to change its intended variable, or that the measurement process is unreliable. The research state should encode these possibilities before execution so that the result triggers a meaningful update rather than post hoc reinterpretation.

After an external observation $o$, revision updates whichever objects are implicated:

\begin{equation}
(P',R',H',X',U')
=
\operatorname{Revise}(P,R,H,X,U;o),
\label{eq:evidence-grounded-revision}
\end{equation}

where $X$ denotes the intervention plan and $U$ local uncertainty. Raw observations remain separate from their interpretations so that old evidence can be reanalyzed after a representation change. Provenance, calibration, protocol fidelity, leakage, and replication determine whether $o$ is eligible to support a scientific update \citep{nosek2015top,wilkinson2016fair}.

Failure attribution is part of the operator. Theory failure, measurement error, implementation error, protocol deviation, hidden confounding, random noise, environmental shift, and simulator misspecification imply different next actions. This prevents both premature abandonment and ad hoc protection of a favored explanation.

\subsection{Episode Output and Candidate Discovery Lessons}
\label{subsec:learning-from-discovery}

A completed episode yields two different artifacts. The first is domain knowledge: supported observations, mechanisms, predictions, validity boundaries, justified negative results, or a defensible conclusion that the current question is not testable. The second is a set of candidate lessons about the research process itself.

A candidate lesson can state that a formulation was too broad, a representation omitted a variable, an intervention was non-identifying, or a replication step prevented a false update. The trajectory alone does not validate the lesson. Success may depend on privileged information or an unrecorded human correction; failure may come from execution rather than from the decision that preceded it. The Discovery Process therefore preserves states, alternatives, actions, observations, and provenance without immediately converting retrospective explanations into reusable rules.

Zetema, introduced next, provides the system mechanism for maintaining this state over long horizons and deciding which candidate lessons are allowed to influence future discovery.

\section{Zetema: A System Instantiation of Discovery Intelligence}
\label{sec:zetema}

\begin{figure}[!htbp]
  \centering
  \includegraphics[width=\linewidth]{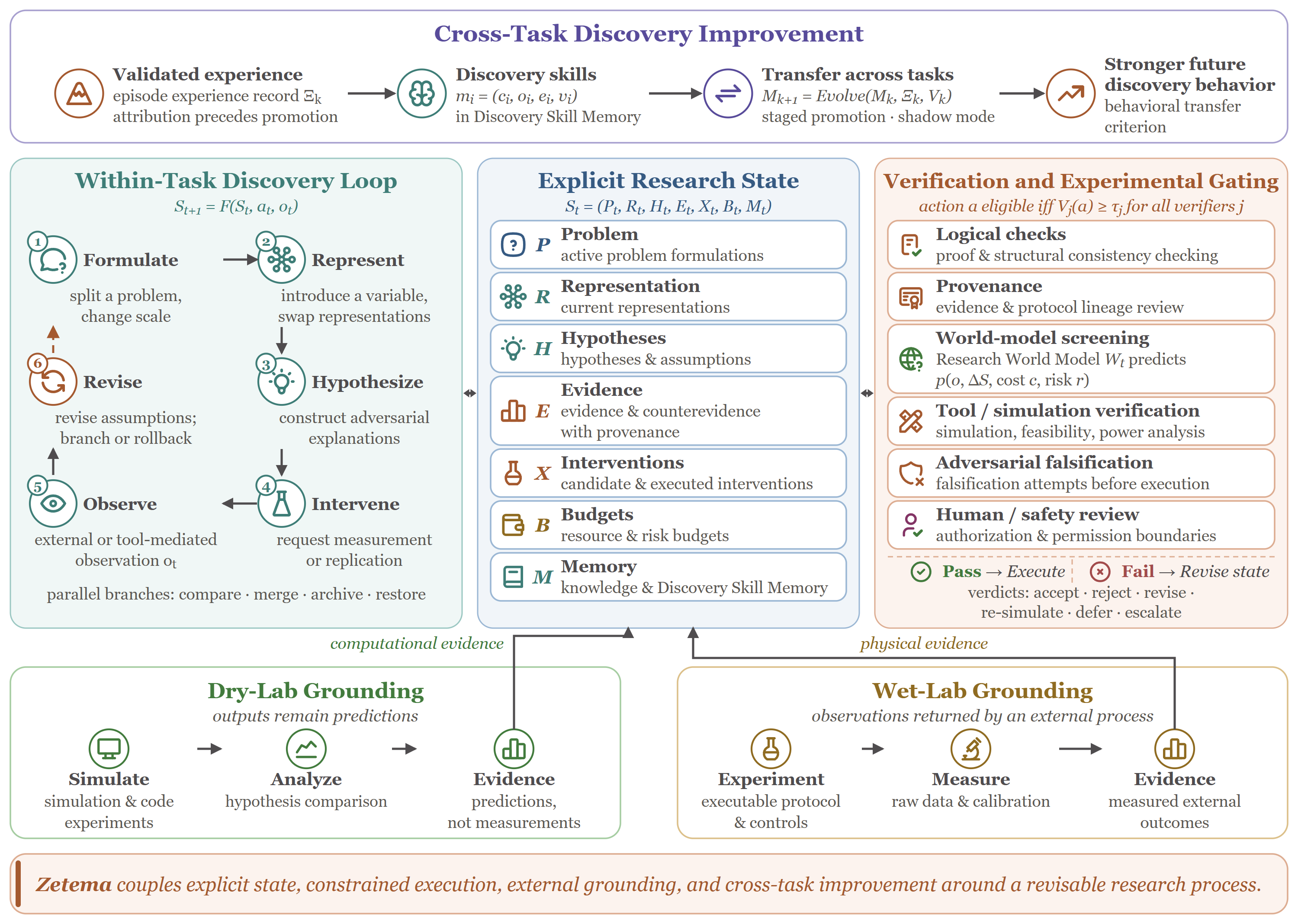}
\caption{
\textbf{Zetema: a system instantiation of Discovery Intelligence.}
Zetema organizes discovery around an explicit and revisable research state that couples a within-task discovery loop with verification and experimental gating. Dry-Lab and Wet-Lab interfaces return computational and physical evidence to the shared state, while validated episode-level experience is attributed, consolidated into Discovery Skills, and transferred across tasks to improve future discovery behavior.
}
\label{fig:zetema}
\end{figure}

We instantiate the DFM framework with \emph{Zetema}, a system organization that couples within-task research-state revision, evidence-based action gating, external experimentation, and cross-task Discovery Skill evolution. Zetema specifies functional interfaces rather than a mandatory neural architecture: foundation models, tools, simulators, verifiers, laboratories, and human researchers can implement different parts of the same organization. Figure~\ref{fig:zetema} gives the resulting data and control flow.

Zetema is built around one constraint: a scientifically consequential operation must leave an inspectable state transition. Problems, representations, hypotheses, evidence, interventions, uncertainties, budgets, and reusable experience therefore remain distinguishable even when their internal implementation is neural or unstructured. This state makes revision, branching, attribution, and later training possible.

\subsection{Within-Task Research-State Dynamics}
\label{subsec:within-task-discovery}

At step $t$, Zetema maintains a research state $S_t$, selects a research operation $a_t$, receives an external or tool-mediated observation $o_t$, and applies

\begin{equation}
S_{t+1}=\mathcal{F}(S_t,a_t,o_t).
\label{eq:zetema-state-update}
\end{equation}

The action space includes ordinary tool use---searching literature, executing code, calling a simulator, requesting a measurement---and operations on the research program itself. The system can split a problem, introduce a variable, change scale, replace a representation, construct an adversarial explanation, revise an assumption, request replication, or terminate a branch.

This distinction changes how long-horizon reasoning is organized. When several interpretations of a failure remain plausible, Zetema does not need to compress them into one uncertain narrative. It can maintain parallel branches, associate each branch with a diagnostic action, and compare the resulting evidence. A residual, for example, can be tracked simultaneously as a possible missing variable, dataset shift, or implementation error until an intervention separates those accounts.

Branches are first-class state objects. They can be compared, merged when their assumptions become compatible, archived when their expected value falls, or restored after new evidence changes their status. Termination is also explicit: a branch can stop because no feasible intervention is identifying, the original effect fails to replicate, the risk exceeds the expected value, or another research opportunity becomes more informative. A valid episode therefore need not end in a positive discovery claim.

\subsection{Explicit Research State and Attribution}
\label{subsec:explicit-research-state}

We instantiate the state as

\begin{equation}
S_t=(P_t,R_t,H_t,E_t,X_t,B_t,\mathcal{M}_t),
\label{eq:explicit-research-state}
\end{equation}

where $P_t$ contains active problem formulations, $R_t$ the current representations, $H_t$ hypotheses and assumptions, $E_t$ evidence and counterevidence with provenance, $X_t$ candidate and executed interventions, $B_t$ resource and risk budgets, and $\mathcal{M}_t$ the knowledge and Discovery Skill Memory available to the episode.

These fields form a linked record rather than separate notes. A hypothesis points to the problem and representation under which it is meaningful. An observation points to the intervention and protocol that produced it. A revision points to the evidence that triggered the change. Human edits, verifier rejections, execution deviations, and permission boundaries are retained because they affect both scientific attribution and the later training signal.

We also type operations by scientific role. A representation revision is not stored as another hypothesis; a predicted outcome is not stored as an observed measurement; a rejected branch is not deleted. Local uncertainty is attached to the object it concerns. The system can therefore trust a measurement while remaining uncertain about its mechanism, or accept an explanation inside one regime without extending it beyond its validity range.

The implementation can use structured text, graphs, databases, programs, equations, or hybrid representations. Zetema requires only that the research objects and their revision relations be recoverable well enough to support comparison, validation, rollback, and attribution.

\subsection{Discovery Skill Memory}
\label{subsec:discovery-skills}

Cross-task improvement requires a representation of reusable research operations rather than an archive of successful trajectories. Thought templates provide a reasoning-level precedent: Buffer of Thoughts distills and reuses methods across tasks \citep{yang2024buffer}, while ReasonFlux learns to select and organize templates into hierarchical reasoning trajectories \citep{yang2025reasonflux}. Discovery Skills extend this procedural view to research operations with explicit state-dependent triggers, expected effects, and external validation requirements. Zetema stores a candidate Discovery Skill as

\begin{equation}
m_i=(c_i,o_i,e_i,v_i),
\label{eq:discovery-skill}
\end{equation}

where $c_i$ is a trigger over research-state conditions, $o_i$ the operation to consider, $e_i$ its expected effect, and $v_i$ the evidence required to validate that effect.

Triggers are structural. A skill can activate when several hypotheses make the same prediction under all current measurements, when errors cluster at a regime boundary, when repeated interventions remain non-identifying, or when a theory accumulates local exceptions. The corresponding operation may request a new measurement, search for a hidden variable, change scale, construct a boundary test, replicate an effect, or stop an unproductive direction.

The expected effect prevents memory from becoming a collection of generic advice. A skill should state what is expected to change---for example, separating hypotheses, reducing uncertainty over a hidden variable, lowering experimental cost, or exposing measurement failure. Its validation record stores supporting episodes, counterexamples, uncertainty, and known domain restrictions. A retrieved skill can therefore be invoked provisionally and weakened when its expected effect does not appear.

This formulation separates three different kinds of accumulation: facts about the world, task-specific episodic memory, and operations intended to improve future discovery behavior. Only the last category constitutes Discovery Skill improvement.

\subsection{Research World Model and Experimental Gating}
\label{subsec:world-model-verification}

Open-ended generation produces many fluent but redundant, infeasible, unsafe, or empirically indistinguishable candidates. Zetema therefore inserts a verification and gating layer before actions acquire substantial cost or consequence. A Research World Model $W_t$ predicts possible observations, state changes, costs, and risks:

\begin{equation}
W_t:(S_t,a)\longrightarrow p(o,\Delta S,c,r\mid S_t,a).
\label{eq:research-world-model}
\end{equation}

$W_t$ can combine learned predictors, causal or mechanistic models, simulators, digital twins, formal tools, and ensembles. Its purpose is not to certify truth. It screens counterfactual consequences: which hypotheses would separate, which state objects might change, how much the action costs, and where the model itself is uncertain.

Because screening changes which research directions receive resources, world-model error has a selection effect. A conservative model can suppress unusual but valid interventions; an overconfident model can repeatedly prefer actions implied by its own misspecification. Zetema therefore retains model disagreement, calibration, validity ranges, and out-of-distribution signals. High uncertainty can trigger a bounded pilot rather than automatic rejection, and repeated disagreement between predictions and external outcomes updates the screening model itself.

Verification is adaptive to the action. A formal claim may proceed directly to proof checking. A physical intervention can require structural checks, provenance review, power analysis, simulation, adversarial falsification, feasibility checks, and human authorization. We represent eligibility as

\begin{equation}
a\in\mathcal{A}_{\mathrm{eligible}}
\iff
V_j(a)\geq\tau_j
\quad\forall j\in\mathcal{J},
\label{eq:experimental-gating}
\end{equation}

where each $V_j$ can be a structured verifier, an uncertainty-aware model, a formal constraint, or a human decision. Thresholds depend on the action: speculative hypotheses can enter the state under uncertainty, whereas costly, irreversible, or high-risk interventions require stronger evidence and permission.

The gate can accept, reject, request revision, request another simulation, defer, or escalate. Among eligible actions, the system can retain a portfolio rather than collapsing to one score: low-risk actions for efficient progress, high-information actions for resolving a central uncertainty, and bounded high-uncertainty actions that test a potentially transformative alternative.

\subsection{Validated Cross-Task Update}
\label{subsec:cross-task-improvement}

After episode $k$, Zetema receives an experience record $\Xi_k$ containing state transitions, successful and failed branches, external outcomes, human interventions, and validation results. Long-term memory evolves as

\begin{equation}
\mathcal{M}_{k+1}
=
\operatorname{Evolve}(\mathcal{M}_k,\Xi_k,\mathcal{V}_k),
\label{eq:cross-task-memory-update}
\end{equation}

where $\mathcal{V}_k$ denotes the validation applied to candidate updates. The update can create, refine, specialize, compose, weaken, or delete a skill.

Attribution precedes promotion. A failed experiment can originate from problem selection, representation, intervention design, execution, or validation. A successful trajectory may depend on an expert-supplied variable or privileged data. Zetema uses the linked research state to identify the state--operation relation actually supported by the episode instead of turning the final retrospective narrative into a general rule.

Recuris provides a concrete precedent in long-horizon agent harnesses: working memory guides experiential skill selection, while execution evidence supports localized, validation-gated memory updates \citep{yu2026recuris}. For discovery, this pattern additionally requires attributing updates to externally grounded research outcomes and testing whether they improve subsequent research decisions.

Promotion can be staged. A candidate skill is first tested on archived trajectories or counterfactual replay, then used in shadow mode on new tasks, and only later allowed to affect active research decisions. Held-out tasks, alternative environments, ablations, expert review, and independent replication test the claimed effect. Skills remain versioned and reversible; negative evidence narrows their trigger or removes them.

The criterion is behavioral transfer. Cross-task improvement is established only when future discovery decisions improve after controlling for additional domain facts, near-duplicate retrieval, prompt reuse, and extra compute.

\subsection{Dry-Lab and Wet-Lab Grounding}
\label{subsec:physical-experiments}

Zetema separates generated expectations from observed outcomes by coupling a Dry-Lab Discovery Loop to a Wet-Lab Grounding Loop. The same interface also covers non-physical environments such as code execution and formal systems; the distinction is whether the observation is generated internally or returned by an external process.

The Dry-Lab loop performs literature synthesis, data analysis, code experiments, simulation, hypothesis comparison, protocol drafting, power analysis, outcome prediction, and failure-mode analysis. Its outputs remain predictions. Expected effect sizes and simulated observations do not enter $E_t$ with the same epistemic status as measurements.

When a physical experiment is required, a selected scientific intention is translated into an executable protocol with variables, controls, samples, measurements, expected outcomes, stopping conditions, and checks for contamination or manipulation failure. Execution records deviations instead of assuming perfect compliance. Measurements retain raw data, calibration state, batch and instrument effects, missing observations, replicate consistency, and sample provenance.

Protocol translation is itself diagnostic. An intervention that appears identifying in abstract form can become impossible under available instrument resolution, sample size, manipulation range, or safety constraints. In that case the Wet-Lab interface returns a formulation or representation failure to the research state rather than simply a binary feasibility rejection.

Zetema therefore closes the method loop without assigning discovery to one component. Foundation models propose and revise research objects; tools and environments return consequences; validators constrain promotion and action; human researchers contribute domain judgment, authorization, criticism, execution, and replication. The capability claim belongs to the declared integrated system and its recorded state transitions. Importantly, this form of Dry-Lab/Wet-Lab coupling is not only hypothetical: the following case study shows a real experimental loop in which physical biological outcomes revise subsequent discovery decisions.

\subsection{Empirical Case Study: A Real Dry-Lab/Wet-Lab Discovery Loop}
\label{subsec:galileo-case}

\begin{figure}[!htbp]
    \centering
    \includegraphics[width=0.99\linewidth]{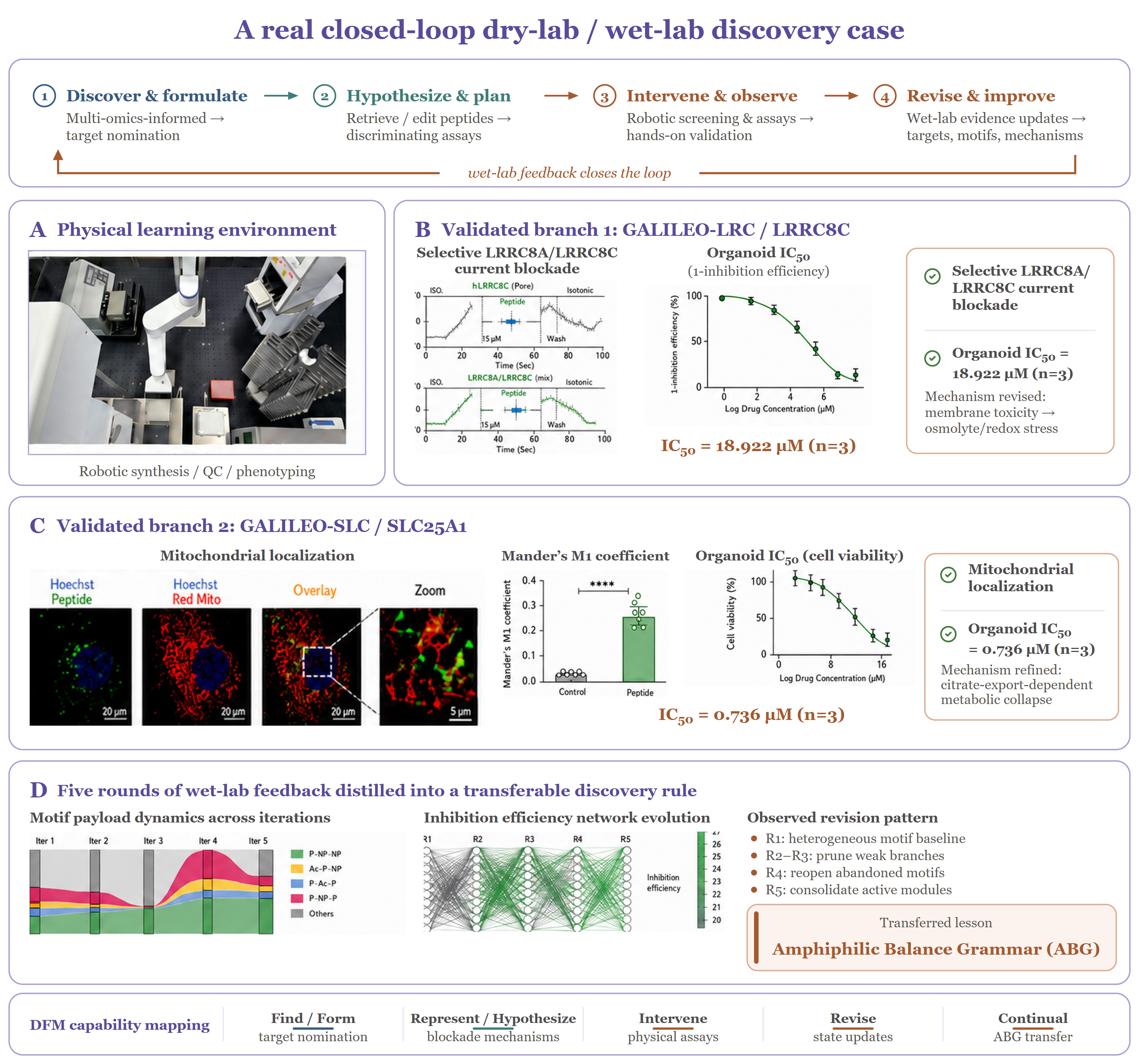}
    \caption{
    \textbf{A real closed-loop Dry-Lab/Wet-Lab discovery case.}
    GALILEO provides an empirical example of how several Discovery Foundation Model operations can be instantiated in a physically grounded scientific workflow. The top panel summarizes the closed loop from discovery and formulation, through hypothesis and intervention design, to physical observation and evidence-grounded revision.
    \textbf{(A)} The physical learning environment couples robotic synthesis, quality control, liquid handling, imaging, and plate-based phenotyping to the discovery process.
    \textbf{(B)} One experimentally validated branch produces GALILEO-LRC, supported by selective LRRC8A/LRRC8C current blockade and organoid activity with an $\mathrm{IC}_{50}$ of $18.922\,\mu\mathrm{M}$.
    \textbf{(C)} A second branch produces GALILEO-SLC, supported by mitochondrial localization and organoid activity with an $\mathrm{IC}_{50}$ of $0.736\,\mu\mathrm{M}$.
    \textbf{(D)} Across five rounds of wet-laboratory feedback, weak motif organizations are pruned, previously abandoned organizations are reopened, and productive modules are consolidated, yielding the Amphiphilic Balance Grammar (ABG) as a reusable design rule.
    The figure therefore represents a real experimentally grounded closed loop rather than a hypothetical workflow: physical evidence changes subsequent scientific actions and contributes to transferable discovery behavior. 
    }
    \label{fig:galileo-case}
\end{figure}

The Dry-Lab/Wet-Lab interface above is not only a conceptual organization. A concrete instance of several of these operations already appears in a real therapeutic-discovery setting. Figure~\ref{fig:galileo-case} summarizes GALILEO, an embodied AI-scientist system for therapeutic peptide discovery in dynamic membrane systems (\href{https://doi.org/10.64898/2026.06.10.731360}{Jiang et al., 2026}). Rather than treating the wet laboratory as a terminal validation stage, GALILEO places experimentally returned biological outcomes inside the iterative decision loop: candidate interventions are proposed, physically executed, measured, and used to revise subsequent target beliefs, molecular-design policies, assay choices, and mechanism hypotheses.

\paragraph{A physical environment as part of the learning loop.}
Panel~A makes explicit a distinction that is easy to obscure in purely computational scientific agents. The environment is not only a source of additional context or a tool endpoint; it returns consequences that the model system does not control. GALILEO couples a cognitive discovery layer to robotic synthesis, quality control, liquid handling, imaging, and plate-based phenotyping. Candidate peptides are retrieved from a clinically informed peptide prior and locally edited, after which physical execution produces observations that can support, weaken, or redirect the active research branch. In the terminology of Section~\ref{subsec:physical-experiments}, the resulting measurements enter the evidence state as externally returned observations rather than as model-generated expectations.

This distinction matters because physical feedback can alter the \emph{research program}, not only its final score. In the reported GALILEO loop, plate-based phenotypes and orthogonal validation results were used to update target-branch belief; viability, morphology, solubility, and quality-control outcomes changed motif-policy weights; ambiguous phenotypes triggered new assays; and the evolving hypothesis board changed which mechanisms and interventions were pursued next. The laboratory therefore functions as a capability-forming environment in the stronger sense used throughout this paper: action produces evidence, and evidence changes subsequent scientific decisions.

\paragraph{Externally grounded discovery branches.}
Panels~B and~C illustrate two distinct branches in which computational proposals were subjected to orthogonal physical tests. For the LRRC8C branch, GALILEO-LRC was evaluated through whole-cell electrophysiology and showed selective inhibition of LRRC8A/LRRC8C currents relative to other tested LRRC8 heteromers. Patient-derived organoid experiments provided a second level of evidence, with an experimentally measured $\mathrm{IC}_{50}$ of $18.922\,\mu\mathrm{M}$ in the LRRC8C-high organoid setting. Subsequent metabolomic, transcriptomic, perturbational, and immune-context experiments further shifted the working explanation from generic membrane toxicity toward an LRRC8C-linked osmolyte/redox-stress and innate-signaling mechanism.

The SLC25A1 branch followed a different evidential route. GALILEO-SLC was shown to localize to the mitochondrial compartment in which SLC25A1 operates, and dose-response experiments in SLC25A1-high patient-derived organoids yielded an $\mathrm{IC}_{50}$ of $0.736\,\mu\mathrm{M}$. Target silencing, acetate rescue, mitochondrial measurements, metabolomics, and extracellular-acidification assays then progressively refined the working explanation toward a citrate-export-dependent metabolic-collapse mechanism. These branches are important for the DFM framework not because they merely produce successful candidates, but because heterogeneous external observations support different updates to the research state: target confidence, mechanism, intervention choice, and the next measurement can all change as evidence accumulates.

\paragraph{From repeated feedback to a transferable discovery rule.}
Panel~D provides the strongest connection to evidence-grounded revision and continual discovery improvement. Across five autonomous peptide-optimization rounds, the trajectory was not a monotonic search toward one increasingly dominant candidate family. The first round maintained a heterogeneous motif baseline; rounds~2--3 removed weakly supported branches and contracted the active motif payload; round~4 reopened previously abandoned acidic, polar, and nonpolar organizations under revised pore-engagement hypotheses; and round~5 consolidated recovered motif organizations into shared active modules.

This pattern is qualitatively different from static candidate ranking. A rejected branch is not necessarily forgotten permanently, and a previously weak design operation can become useful after the research state changes. The relevant object of learning is therefore not only ``which peptide worked,'' but \emph{under which research-state conditions a particular molecular organization should be generated, rejected, recovered, or tested again}. The resulting Amphiphilic Balance Grammar (ABG) summarizes this accumulated physical feedback into a reusable intervention prior. In the original study, motifs derived from the physically optimized branches were further transferred into external peptide-design workflows, improving the geometry and energetic profiles of their outputs. This provides an empirical example of the transition from episode-level experience to a more portable discovery operation.

\paragraph{Mapping the case to DFM capabilities.}
The correspondence to the DFM capability decomposition is not one-to-one at the level of software modules, but it is operational at the level of research decisions. Target nomination and translational-gap analysis instantiate aspects of $C_{\mathrm{find}}$ and $C_{\mathrm{form}}$; construction of target-specific blockade mechanisms and sequence organizations engages $C_{\mathrm{repr}}$ and $C_{\mathrm{hyp}}$; robotic and hands-on assays instantiate $C_{\mathrm{int}}$; updates to target belief, motif policy, assay selection, and mechanism hypotheses instantiate $C_{\mathrm{rev}}$; and consolidation of experimentally supported motif structure into ABG provides a concrete instance of experience influencing future discovery behavior, corresponding to the direction targeted by $C_{\mathrm{cont}}$.

The case also illustrates why a DFM is defined as a \emph{model system} rather than as a single autonomous model. Not every experimentally consequential operation in GALILEO is executed by the robotic platform. Synthesis, quality control, liquid handling, imaging, and plate-based phenotyping are integrated into the automated loop, whereas specialized assays including electrophysiology, metabolic profiling, patient-derived tissue experiments, and in-vivo validation can remain agent-specified but human-executed. These interventions should remain explicitly represented in the system boundary and provenance record rather than being attributed to the model alone. The relevant capability claim belongs to the coupled system of policies, tools, experimental environments, validators, and human execution described by Equation~\ref{eq:dfm-system}.

\paragraph{What this case establishes---and what it does not.}
GALILEO should not be interpreted as evidence that the full general-purpose DFM problem has already been solved. Its problem domain, molecular modality, experimental interfaces, and scientific objectives remain substantially structured, and broader cross-domain discovery transfer is still an open problem. Its importance here is narrower but concrete: it demonstrates a \emph{real physical discovery loop} in which externally generated biological evidence is not merely used to validate a final model proposal. Instead, that evidence changes subsequent scientific actions, can reopen or terminate research branches, modifies the policy used to construct later interventions, and can ultimately be distilled into a reusable discovery rule.

This empirical distinction is central to Discovery Intelligence. A conceptual agent loop can always be drawn as
\[
\text{hypothesis} \rightarrow \text{experiment} \rightarrow \text{result},
\]
but a grounded discovery loop requires the stronger transition
\[
S_t
\xrightarrow{\;a_t\;}
o_t
\xrightarrow{\;\text{external evidence}\;}
S_{t+1},
\]
where $S_{t+1}$ can change not only the preferred answer, but the problem interpretation, representation, mechanism, intervention strategy, evaluator, or reusable discovery operation. The GALILEO case shows that such evidence-driven state revision can already be realized in a real Dry-Lab/Wet-Lab scientific workflow. Capability formation, discussed next, asks how these state-conditioned research operations can be learned and generalized systematically.

\section{Capability Formation: Training Discovery Operations}
\label{sec:capability-formation}

Zetema specifies how discovery is organized; capability formation specifies how the underlying research policy is learned. We formulate training around state-conditioned decisions rather than final scientific answers. A training example is useful when it exposes which research operation was available, why it was selected, what external consequence followed, and how that consequence changed the state.

\begin{figure}[!htbp]
  \centering
  \includegraphics[width=\linewidth]{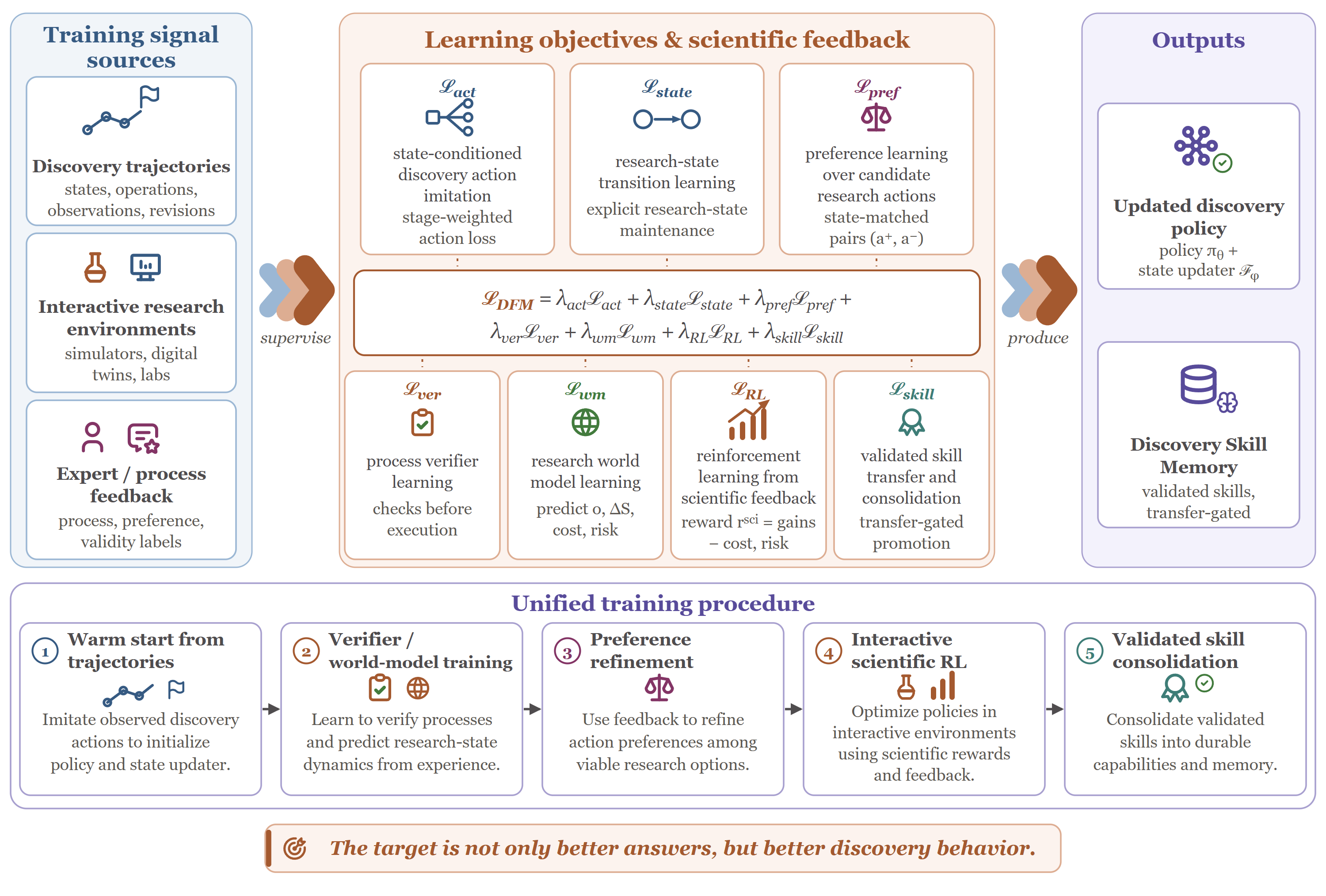}
\caption{
\textbf{Capability formation as multi-signal optimization of discovery behavior.}
Training combines state-conditioned discovery-action imitation, research-state transition learning, preference optimization, process verification, Research World Model learning, reinforcement learning from scientific feedback, and validated Discovery Skill transfer. These objectives form the compositional training objective $\mathcal{L}_{\mathrm{DFM}}$ and can be optimized through a staged procedure from trajectory warm-start and verifier/world-model learning to preference refinement, interactive scientific reinforcement learning, and validated skill consolidation.
}
\label{fig:capability-formation}
\end{figure}

Figure~\ref{fig:capability-formation} summarizes the training loop.

\subsection{Training Data and Interactive Research Environments}
\label{subsec:discovery-trajectories}

A discovery trajectory records a sequence of research states, available alternatives, selected operations, external observations, and revisions. The transitions are more important than a polished account of the final result. Useful data include why an observation was treated as an unknown, which formulations were rejected, what representation change made an intervention possible, which assumptions separated the explanations, and what later evidence showed that a decision was productive.

Success-only trajectories are insufficient. Training data should retain malformed questions, missing variables, misleading representations, non-identifying interventions, failed replications, contradictory evidence, abandoned branches, and attribution errors. These failures teach different policies only when their source is visible. An inconclusive experiment caused by low statistical power should not supervise the same update as an experiment that was well executed but non-identifying.

Historical artifacts provide partial signals. Laboratory notebooks, code histories, preregistrations, peer review, rebuttals, failed replications, and revised manuscripts expose changes in the research state but often omit alternatives or rationalize decisions retrospectively. Synthetic and simulated trajectories can provide latent mechanisms, counterfactual actions, and known failure sources, but inherit the assumptions of their generator. We therefore treat data-source provenance as part of the trajectory rather than flattening all examples into one demonstration format.

Interactive environments complement static trajectories by making the model choose what to observe or test next. Useful environments include causal worlds, formal systems, codebases, machine-learning experiments, scientific simulators, digital twins, robotic platforms, and human-mediated laboratories \citep{gandhi2025boxinggym,majumder2025discoverybench,huang2024mlagentbench,hafner2025dreamer}. They should hide scientifically meaningful structure while retaining external outcomes that make research decisions evaluable.

A discovery environment should not reveal the correct question, variables, action space, and success criterion simultaneously. Instead it can contain hidden variables, observationally equivalent mechanisms, noisy evidence, evaluator mismatch, or cases in which the supplied representation is deliberately inadequate. The model then chooses an intervention and receives an observation it did not write itself. In simulators, unchosen actions can also be evaluated counterfactually, providing denser credit for intervention selection.

For empirical domains, training authority can increase from static data and executable code to simulation, digital twins, shadow-mode protocol planning, and supervised physical experiments. The curriculum should vary the type of uncertainty as well as realism so that the model does not memorize one canonical discovery workflow. Difficulty-aligned agent--environment co-evolution, as explored in GenEnv \citep{guo2025genenv}, provides a mechanism for adapting training tasks to current capabilities. In discovery settings, such adaptation should vary hidden mechanisms and sources of uncertainty while retaining independently evaluable outcomes.

\subsection{Learning Objectives and Scientific Feedback}
\label{subsec:process-supervision}

We parameterize the trainable parts of the discovery system as a research policy $\pi_{\theta}$, a state updater $\mathcal{F}_{\phi}$, a Research World Model $W_{\omega}$, a process verifier $V_{\psi}$, and a Discovery Skill retriever or selector $q_{\eta}$. These components need not be separate neural networks; the notation identifies the functions receiving distinct training signals. A training record contains a research state $S_t$, an available or selected operation $a_t$, an externally returned observation $o_t$, the resulting state $S_{t+1}$, and any process, preference, validity, cost, or transfer labels available for that transition.

\paragraph{State-conditioned discovery imitation.}
Discovery trajectories provide supervision over which research operation is appropriate under a particular state. Let $z_t\in\{\mathrm{find},\mathrm{form},\mathrm{repr},\mathrm{hyp},\mathrm{int},\mathrm{rev}\}$ denote the scientific role of a transition. We use a stage-weighted action loss

\begin{equation}
\mathcal{L}_{\mathrm{act}}
=
-\mathbb{E}_{\Xi\sim\mathcal{D}_{\mathrm{traj}}}
\left[
\sum_t \alpha_{z_t}
\log \pi_{\theta}(a_t^{*}\mid S_t,\mathcal{M}_t)
\right],
\label{eq:discovery-action-loss}
\end{equation}

where $a_t^{*}$ is a demonstrated or validated research operation and $\alpha_{z_t}$ prevents abundant operation types from dominating rarer discovery transitions. The same trajectory supervises explicit research-state maintenance:

\begin{equation}
\mathcal{L}_{\mathrm{state}}
=
-\mathbb{E}_{\Xi\sim\mathcal{D}_{\mathrm{traj}}}
\left[
\sum_t
\log p_{\phi}
\left(
S_{t+1}\mid S_t,a_t,o_t
\right)
\right].
\label{eq:research-state-loss}
\end{equation}

This term trains the system to preserve which problem, representation, hypothesis, evidence item, intervention, or budget changed after an observation rather than compressing the transition into an undifferentiated narrative.

\paragraph{Preference learning over research operations.}
When two actions are compared under the same research state and budget, preference optimization directly trains the policy to select the stronger research decision. For a preferred action $a^{+}$ and a rejected action $a^{-}$, one possible pairwise objective is

\begin{equation}
\begin{aligned}
\mathcal{L}_{\mathrm{pref}}
={}&-\mathbb{E}_{(S,a^{+},a^{-})\sim\mathcal{D}_{\mathrm{pref}}}
\Bigg[
\log \sigma\Bigg(
\beta\Big[
\log\frac{\pi_{\theta}(a^{+}\mid S)}{\pi_{\mathrm{ref}}(a^{+}\mid S)}
-\log\frac{\pi_{\theta}(a^{-}\mid S)}{\pi_{\mathrm{ref}}(a^{-}\mid S)}
\Big]
\Bigg)
\Bigg],
\end{aligned}
\label{eq:discovery-preference-loss}
\end{equation}

where preferences can reflect testability, information gain, evidential support, cost, robustness, or risk. Conditioning both candidates on the same $S$ is important: an action such as replication or high-variance intervention is not globally good or bad, but appropriate only under particular research conditions.

\paragraph{Process verification.}
The verifier estimates whether a proposed transition satisfies the criteria that apply before execution or promotion. Let $j\in\mathcal{J}$ index criteria such as structural validity, evidential support, discriminative value, feasibility, and safety, with labels $y_t^{(j)}\in\{0,1\}$. A multi-criterion verifier can be trained with

\begin{equation}
\mathcal{L}_{\mathrm{ver}}
=
-\mathbb{E}
\left[
\sum_{t}
\sum_{j\in\mathcal{J}}
\left(
 y_t^{(j)}\log V_{\psi}^{(j)}(S_t,a_t)
+
(1-y_t^{(j)})\log(1-V_{\psi}^{(j)}(S_t,a_t))
\right)
\right].
\label{eq:process-verifier-loss}
\end{equation}

The verifier is not a replacement for external evidence. Its role is to learn reusable checks that filter malformed or unsupported transitions before more expensive interaction.

\paragraph{Research World Model learning.}
For actions that interact with an environment, the Research World Model predicts possible observations, research-state changes, costs, and risks. Using the notation from Equation~\ref{eq:research-world-model}, a generic negative log-likelihood objective is

\begin{equation}
\mathcal{L}_{\mathrm{wm}}
=
-\mathbb{E}_{(S_t,a_t,o_t,\Delta S_t,c_t,\rho_t)\sim\mathcal{D}_{\mathrm{int}}}
\left[
\log W_{\omega}
\left(
 o_t,\Delta S_t,c_t,\rho_t
 \mid S_t,a_t
\right)
\right],
\label{eq:research-world-model-loss}
\end{equation}

where $\rho_t$ denotes action risk. In practice, the joint likelihood can be decomposed into modality-specific prediction, calibration, cost, or risk losses. Keeping the objective at this functional level allows the same formulation to cover formal tools, learned simulators, mechanistic models, and empirical predictors.

\paragraph{Reinforcement learning from scientific feedback.}
Interactive episodes provide feedback that is not available from retrospective demonstrations. We decompose a per-step scientific reward as

\begin{equation}
\begin{aligned}
r_t^{\mathrm{sci}}
={}&
 w_K\,\delta K_t
+w_I\,\mathrm{IG}_t
+w_R\,q_t^{\mathrm{rev}}
+w_V\,q_t^{\mathrm{val}}
-w_C\,c_t
-w_{\rho}\,\rho_t,
\end{aligned}
\label{eq:scientific-step-reward}
\end{equation}

where $\delta K_t$ denotes externally supported knowledge progress, $\mathrm{IG}_t$ realized or counterfactually estimated information gain, $q_t^{\mathrm{rev}}$ the quality of evidence-grounded revision, $q_t^{\mathrm{val}}$ validation quality, and $c_t$ and $\rho_t$ the incurred cost and risk. A terminal transfer signal can reward improvement beyond the current episode:

\begin{equation}
R(\Xi)
=
\sum_{t=0}^{T-1}\gamma^t r_t^{\mathrm{sci}}
+
\lambda_T\Delta C_{\mathrm{future}}.
\label{eq:scientific-trajectory-return}
\end{equation}

A generic KL-regularized policy-gradient objective is then

\begin{equation}
\mathcal{L}_{\mathrm{RL}}
=
-\mathbb{E}_{\Xi\sim\pi_{\theta}}
\left[
\sum_t \widehat{A}_t
\log\pi_{\theta}(a_t\mid S_t,\mathcal{M}_t)
\right]
+
\beta_{\mathrm{KL}}
\mathbb{E}_{S_t}
\left[
D_{\mathrm{KL}}
\left(
\pi_{\theta}(\cdot\mid S_t)
\,\|\,
\pi_{\mathrm{ref}}(\cdot\mid S_t)
\right)
\right],
\label{eq:scientific-rl-loss}
\end{equation}

where $\widehat{A}_t$ can be obtained by any suitable long-horizon credit-assignment method. Equation~\ref{eq:scientific-rl-loss} specifies the training signal rather than committing Zetema to PPO, GRPO, or another particular optimizer.

\paragraph{Validated Discovery Skill learning.}
Cross-task improvement requires learning when a reusable Discovery Skill applies. Let $m^{+}$ be a skill whose trigger and expected effect have been validated on held-out or shadow-mode episodes, and let $\mathcal{N}(S)$ contain irrelevant or harmful skills for the same state. A contrastive retrieval objective is

\begin{equation}
\mathcal{L}_{\mathrm{skill}}
=
-\mathbb{E}_{(S,m^{+})}
\left[
\log
\frac{
\exp(q_{\eta}(S,m^{+})/\tau)
}{
\exp(q_{\eta}(S,m^{+})/\tau)
+
\sum_{m^{-}\in\mathcal{N}(S)}
\exp(q_{\eta}(S,m^{-})/\tau)
}
\right].
\label{eq:discovery-skill-loss}
\end{equation}

Only validated skill updates should be treated as positive transfer targets. Counterexamples and negative-transfer cases populate $\mathcal{N}(S)$ so that the model learns when \emph{not} to reuse a previously successful operation.

The objectives can be summarized as

\begin{equation}
\begin{aligned}
\mathcal{L}_{\mathrm{DFM}}
={}&
\lambda_{\mathrm{act}}\mathcal{L}_{\mathrm{act}}
+\lambda_{\mathrm{state}}\mathcal{L}_{\mathrm{state}}
+\lambda_{\mathrm{pref}}\mathcal{L}_{\mathrm{pref}}
+\lambda_{\mathrm{ver}}\mathcal{L}_{\mathrm{ver}}
+\lambda_{\mathrm{wm}}\mathcal{L}_{\mathrm{wm}}
+\lambda_{\mathrm{RL}}\mathcal{L}_{\mathrm{RL}}
+\lambda_{\mathrm{skill}}\mathcal{L}_{\mathrm{skill}}.
\end{aligned}
\label{eq:dfm-training-objective}
\end{equation}

Equation~\ref{eq:dfm-training-objective} is a compositional specification, not a requirement that all terms be optimized simultaneously. A practical implementation can warm-start the research policy and state updater from trajectories, learn verifiers and world models from labeled interaction data, refine decisions with preferences, optimize long-horizon behavior through scientific feedback, and consolidate only those skills that survive held-out transfer tests.

\subsection{Unified Training Procedure}
\label{subsec:unified-training-procedure}

Algorithm~\ref{alg:dfm-training} gives one reference training procedure corresponding to the objectives above. It separates offline capability acquisition from interactive improvement and keeps long-term memory updates behind a transfer-validation gate.

\begin{algorithm}[t]
\caption{Reference Training Procedure for Discovery Foundation Models}
\label{alg:dfm-training}
\begin{algorithmic}[1]
\REQUIRE trajectory data $\mathcal{D}_{\mathrm{traj}}$, preference data $\mathcal{D}_{\mathrm{pref}}$, interactive environments $\mathcal{E}$, validators $\mathcal{V}$, initial memory $\mathcal{M}$
\STATE Initialize policy $\pi_{\theta}$, state updater $\mathcal{F}_{\phi}$, world model $W_{\omega}$, verifier $V_{\psi}$, and skill selector $q_{\eta}$
\STATE \textbf{Offline warm start:} optimize $\mathcal{L}_{\mathrm{act}}+\lambda_{\mathrm{state}}\mathcal{L}_{\mathrm{state}}$ on $\mathcal{D}_{\mathrm{traj}}$
\STATE Train $V_{\psi}$ with $\mathcal{L}_{\mathrm{ver}}$ and $W_{\omega}$ with $\mathcal{L}_{\mathrm{wm}}$ from labeled or replayed interactions
\STATE Refine $\pi_{\theta}$ on state-matched action pairs using $\mathcal{L}_{\mathrm{pref}}$
\FOR{each interactive discovery episode}
    \STATE Initialize research state $S_0$ from the partially understood world and current memory $\mathcal{M}$
    \FOR{$t=0,\ldots,T-1$}
        \STATE Propose candidate research operations $\mathcal{A}_t\sim\pi_{\theta}(\cdot\mid S_t,\mathcal{M})$
        \STATE Predict outcomes, costs, and risks with $W_{\omega}$; score applicable constraints with $V_{\psi}$
        \STATE Gate candidates and select an eligible operation $a_t$
        \STATE Execute $a_t$ in $\mathcal{E}$ or through an authorized tool; observe $o_t$, cost $c_t$, and risk signal $\rho_t$
        \STATE Update $S_{t+1}=\mathcal{F}_{\phi}(S_t,a_t,o_t)$ and compute $r_t^{\mathrm{sci}}$
        \STATE Store $(S_t,a_t,o_t,S_{t+1})$ with provenance, alternatives, and validation outcomes
    \ENDFOR
    \STATE Update $\pi_{\theta}$ using $\mathcal{L}_{\mathrm{RL}}$; update $W_{\omega}$ and $V_{\psi}$ from newly grounded transitions
    \STATE Infer candidate Discovery Skills from attributed successes, failures, and branch contrasts
    \STATE Evaluate candidate skills on held-out, replay, or shadow-mode episodes
    \IF{a candidate improves transfer under matched resources and passes validation}
        \STATE Promote or revise the skill in $\mathcal{M}$ and update $q_{\eta}$ using $\mathcal{L}_{\mathrm{skill}}$
    \ENDIF
\ENDFOR
\ENSURE trained discovery policy and validated Discovery Skill Memory
\end{algorithmic}
\end{algorithm}

The algorithm is intentionally modular. Components can share parameters, and individual training stages can be omitted when the corresponding supervision is unavailable. What is invariant is the supervision structure: research decisions are learned from state-conditioned trajectories, constrained by verifiers and predictive models, improved through external scientific feedback, and allowed to persist across tasks only after transfer validation.

Coupled optimization has precedents in CURE, which co-trains code and unit-test generators through execution feedback \citep{wang2025cure}, and RLAnything, which jointly adapts environments, policies, and reward models \citep{wang2026rlanything}. These approaches motivate coordinated updates to discovery components, subject to the additional requirement that scientific validity remain anchored in external evidence and independent evaluation.

\subsection{Process-Level Scaling and Resource Allocation}
\label{subsec:discovery-scaling}

Test-time scaling usually allocates more computation to reasoning traces, candidates, search branches, or verifier calls under a fixed task \citep{snell2025testtime}. A DFM can scale different parts of the research process. We allocate budget across problem formulation, representation construction, hypothesis search, intervention design, falsification, and verification:

\begin{equation}
B=B_P+B_R+B_H+B_X+B_F+B_V.
\label{eq:discovery-budget}
\end{equation}

The allocation policy is state-dependent. More hypotheses do little when every candidate makes the same prediction under the available observable. More experiments do little when the representation excludes the relevant variable. Replication can dominate novelty when the effect itself is unstable, while independent verification deserves additional budget when the evaluator is suspected of leakage or shortcut exploitation.

The budget includes more than tokens: tool calls, simulator runs, experimental cost, human time, latency, reversibility, and risk all constrain the next action. A learned allocator can use local uncertainty, world-model disagreement, branch value, and expected information gain to identify the current bottleneck. This converts process-level scaling from a fixed recipe into another discovery policy that can itself be trained and evaluated.

Falsification receives an explicit budget because candidate generation and confirmation otherwise dominate compute allocation. The system can search boundary conditions, contradictory datasets, adversarial explanations, and evaluator shortcuts even when these actions reduce the probability of preserving its current leading hypothesis.

\subsection{Continual Skill Learning and Transfer}
\label{subsec:continual-skill-evolution}

Capability formation also trains the update policy that proposes, tests, and consolidates Discovery Skills. Candidate updates can arise from successful branches, failures, or contrasts between two branches that differ in one consequential operation. These contrasts often support cleaner attribution than a retrospective summary of a single trajectory.

Process supervision can label whether the inferred source of success or failure is plausible. Counterfactual replay tests whether the proposed skill would have changed earlier decisions in the claimed direction. Shadow-mode deployment collects evidence on new tasks before the skill affects active research. Promotion is uncertainty-aware and reversible; contradictory triggers, duplicate skills, spurious correlations, benchmark-specific shortcuts, and retrospective rationalizations remain explicit objects of validation.

Transfer is trained through variation in latent research structure rather than only topic similarity. Hidden-variable diagnosis can appear in causal simulation, machine-learning debugging, and experimental measurement. Boundary testing can be instantiated in algorithms, materials, or biological regimes. Non-identifiability can be expressed through different observables and tool interfaces. Surface variation discourages retrieval based only on vocabulary.

Negative cases are equally important. A representation change should not be triggered after every failed hypothesis; replication should not become a universal response to disagreement; information gain should not override feasibility or safety. Learning when \emph{not} to invoke a skill is part of learning its trigger.

The memory update changes the distribution of future experience, so controlled exploration is required. Frequently retrieved skills can crowd out alternatives and create self-confirming evidence. Periodic evaluation without the skill, tasks selected independently of the current memory, conflict detection, and rollback reduce this feedback. Formation can encourage transfer; Section~\ref{sec:evaluation} specifies the matched controls required to establish that transfer actually occurred.

\section{Capability Evaluation: A Process-Centered Protocol}
\label{sec:evaluation}

\begin{figure}[!htbp]
  \centering
  \includegraphics[width=\linewidth]{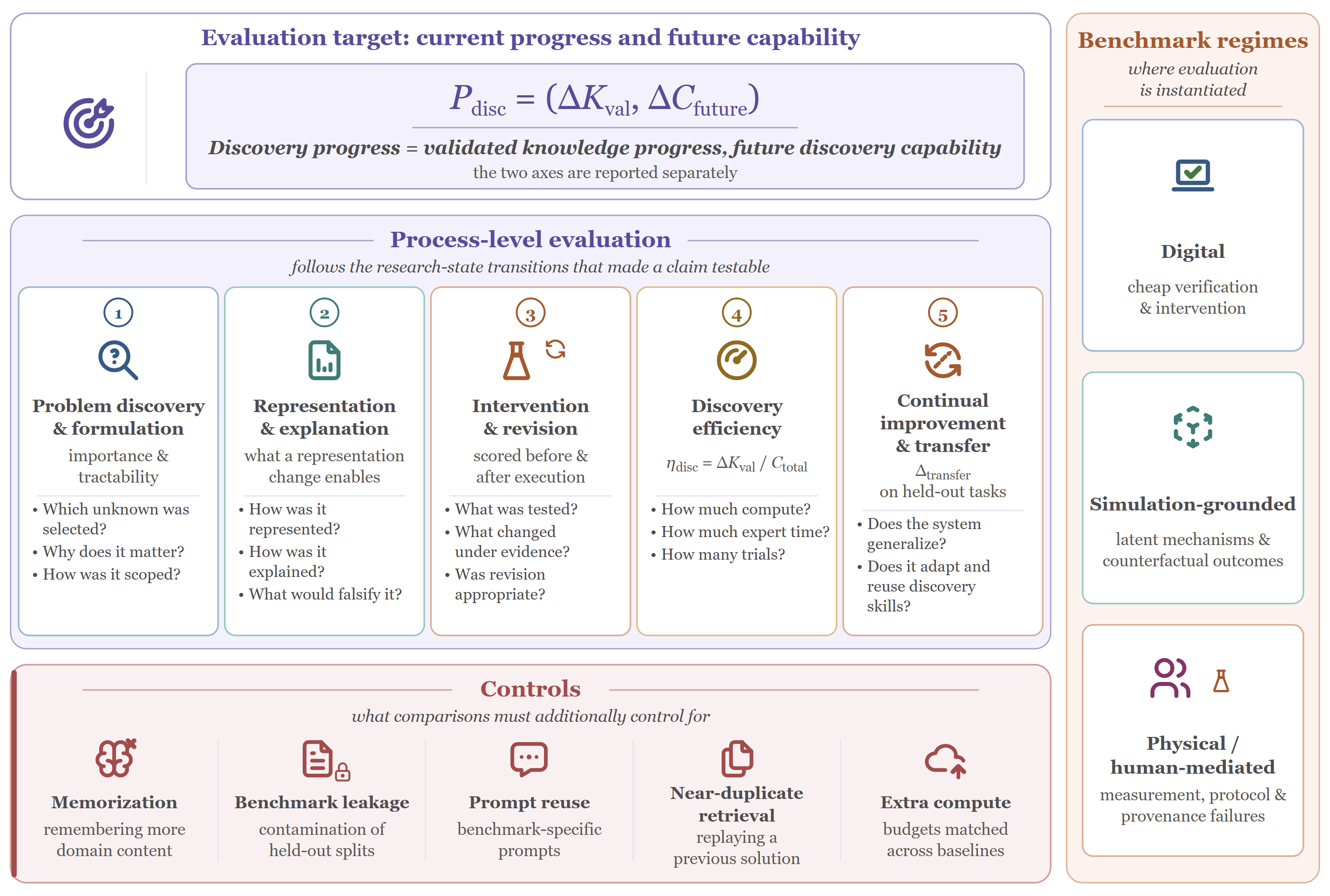}
\caption{
\textbf{A process-centered evaluation protocol for Discovery Foundation Models.}
Discovery progress is evaluated along two complementary axes, externally validated knowledge progress $\Delta K_{\mathrm{val}}$ and improvement in future discovery capability $\Delta C_{\mathrm{future}}$. Process-level evaluation covers problem discovery and formulation, representation and explanation, intervention and revision, discovery efficiency, and continual improvement and transfer. Comparisons additionally control for memorization, benchmark leakage, prompt reuse, near-duplicate retrieval, and additional computation, and can be instantiated across digital, simulation-grounded, and physical or human-mediated environments.
}
\label{fig:capability-evaluation}
\end{figure}

A DFM should not receive discovery credit simply for producing a novel statement, a plausible hypothesis, or a high score under a fixed evaluator. Evaluation follows the research-state transitions that made a claim testable and asks whether experience changes later discovery behavior. Figure~\ref{fig:capability-evaluation} summarizes the protocol.

\subsection{Evaluation Target: Current Progress and Future Capability}
\label{subsec:defining-discovery-progress}

We represent Discovery Progress as

\begin{equation}
\mathcal{P}_{\mathrm{disc}}
=
(\Delta K_{\mathrm{val}},\Delta C_{\mathrm{future}}),
\label{eq:discovery-progress}
\end{equation}

where $\Delta K_{\mathrm{val}}$ is externally validated progress in the current investigation and $\Delta C_{\mathrm{future}}$ is improvement in future discovery behavior. The two axes are reported separately.

$\Delta K_{\mathrm{val}}$ can include recovering a hidden variable, separating regimes, producing a more accurate intervention model, falsifying a mechanism, obtaining a reproducible empirical result, or establishing that a proposed effect does not survive replication. The outcome must be assessed by evidence the proposing system does not control.

$\Delta C_{\mathrm{future}}$ concerns behavior after the episode. The relevant question is whether the system becomes better at recognizing malformed questions, constructing useful representations, choosing discriminating interventions, responding to counterevidence, or allocating resources to the actual bottleneck. Additional factual knowledge alone is not sufficient.

Final outcomes and process profiles are both reported. Two systems can reach the same conclusion while differing in representation quality, number of non-identifying interventions, calibration, or response to contradictory evidence. These differences reveal which discovery operations have actually formed.

\subsection{Stage-Wise Process Evaluation}
\label{subsec:evaluating-problems}

Problem Discovery is evaluated in settings where the research opportunity is not explicitly stated. Inputs can contain observations, literature fragments, failed experiments, inconsistent results, partial goals, and distractor anomalies. Precision matters: a system that launches a research program around every unexplained residual is not demonstrating good problem discovery. Controlled environments can measure whether resolving the selected unknown exposes useful latent structure, while experts assess dimensions such as scientific importance and tractability \citep{majumder2025discoverybench,chen2025scienceagentbench}.

Problem Formulation is stress-tested by supplying overly broad questions, proxy objectives, inappropriate scales, or observables that cannot identify the mechanism. Evaluation checks whether the system repairs the defect and whether the resulting formulation admits informative interventions under the available resources.

Representation evaluation measures what becomes possible after a representation change. Benchmarks can hide a causal variable, mix regimes, provide an incorrect graph, or choose a scale that obscures the dynamics. A proposed representation is credited when it improves held-out prediction, intervention accuracy, regime separation, compression, or transfer relative to the original state. Exact symbol matching is unnecessary when different representations support the same scientific operations \citep{gandhi2025boxinggym}.

Hypotheses are evaluated for mechanism specificity, assumptions, validity range, discriminative predictions, and falsifiers. Multiple explanations should differ in causal structure, dynamics, latent variables, or intervention response rather than surface wording. Withholding intervention outcomes until the system commits to predictions tests whether the explanation had empirical content before seeing the result.

Interventions are evaluated both prospectively and retrospectively. Before execution, the protocol scores expected discrimination, feasibility, cost, risk, power, and probability of an inconclusive outcome. After execution, it measures realized information gain and the state change produced by the observation. Repeated or simulated environments allow the selected action to be compared with counterfactual alternatives so that a lucky outcome is not confused with a strong intervention policy.

Revision tasks provide reliable counterevidence after the system commits to a formulation, representation, and explanation. Some contradictions reflect theory failure; others are generated by instrument error, protocol deviation, hidden confounding, or environment shift. The score therefore targets failure attribution and the appropriateness of the resulting state transition. Repeated tests and limits on unconstrained auxiliary assumptions detect ad hoc protection of a favored theory.

\subsection{Efficiency and Resource-Matched Evaluation}
\label{subsec:discovery-efficiency}

Discovery quality includes how resources are allocated. We use the conceptual efficiency measure

\begin{equation}
\eta_{\mathrm{disc}}
=
\frac{\Delta K_{\mathrm{val}}}{C_{\mathrm{total}}},
\label{eq:discovery-efficiency}
\end{equation}

while reporting the components of $C_{\mathrm{total}}$ separately because compute, money, human time, and risk are not interchangeable.

Relevant records include inference compute, search branches, tool calls, simulator runs, elapsed time, expert time, physical experiments, replication cost, and action risk. More diagnostic measures include cost to the first identifying intervention, number of repeated non-identifying actions, experiments required to eliminate an incorrect mechanism, and regret relative to an expert or oracle policy \citep{huang2024mlagentbench,chan2025mlebench,garikaparthi2026researchgym}.

Resource matching is essential for comparison. A system with persistent memory or a larger verification portfolio may simply receive more compute or more opportunities to inspect the environment. Baselines should therefore match relevant budgets and separately report gains obtained from additional resources. Efficiency is always conditioned on validity and scientific value; selecting trivial unknowns or terminating difficult investigations can make a system appear cheap without making it a better discoverer.

\subsection{Continual Improvement and Transfer}
\label{subsec:evaluating-transfer}

Continual evaluation unfolds over a sequence of episodes. After early episodes update the permitted memory or policy, the revised system is evaluated on held-out tasks whose solutions and surface forms were unavailable during the update. We summarize transfer as

\begin{equation}
\begin{aligned}
\Delta_{\mathrm{transfer}}
={}&\operatorname{DiscPerf}(\mathcal{D}_{\mathrm{unseen}}\mid\mathcal{M}_{1:k})-\operatorname{DiscPerf}(\mathcal{D}_{\mathrm{unseen}}\mid\mathcal{M}_{0}),
\end{aligned}
\label{eq:evaluation-transfer}
\end{equation}

where $\operatorname{DiscPerf}$ aggregates the relevant process metrics under matched resources.

The control set includes no memory, fact-only memory, full-trajectory retrieval, successful-solution retrieval, domain-specific skill memory, and Discovery Skill Memory. Near-duplicate retrieval, benchmark-specific prompts, and extra inference compute are controlled explicitly. This separates transferable research operations from remembering more domain content or replaying a previous solution.

Within-domain transfer changes the research problem while preserving the field. Cross-task transfer changes the task structure while preserving a discovery operation such as hidden-variable diagnosis. Cross-domain transfer changes the surface domain while retaining a structural research challenge. Negative-transfer tasks measure selective retrieval: a memory system that invokes every learned strategy indiscriminately is not improving discovery capability.

Transfer is reported as a curve over episodes rather than a single end point. The curve reveals sample efficiency, saturation, interference with older skills, and whether gains survive increasingly distant mechanisms. Recursive changes to tools, simulators, evaluators, or memory organization face the same requirement but with stronger independence because the update can also alter how later progress is measured.

\subsection{Benchmark Construction and Controls}
\label{subsec:discovery-benchmarks}

Discovery benchmarks should withhold or corrupt the structures the system is expected to construct while retaining objective consequences. Recent benchmarks already expose complementary parts of this design space through data-driven discovery, research workflows, machine-learning experimentation and replication, interactive experiment design, and scenario-grounded scientific reasoning \citep{gandhi2025boxinggym,majumder2025discoverybench,chen2025scienceagentbench,huang2024mlagentbench,chan2025mlebench,starace2025paperbench,song2025sde}.

An instance can contain an incomplete problem, irrelevant observations, a hidden variable, observationally equivalent mechanisms, noisy evidence, an evaluator shortcut, or a representation mismatch. Some tasks should require active intervention; others should only be solvable after rejecting the supplied framing. Partial progress is retained: recovering the representation, rejecting a false anomaly, or selecting the identifying experiment can all receive process-level credit even when the final discovery is incomplete.

Splits should hold out mechanisms and research structures rather than only natural-language topics. Procedural generation, private instances, new interfaces, and expert-authored challenges reduce contamination and near-duplicate retrieval. Digital tasks provide cheap verification and intervention; simulators expose latent mechanisms and counterfactual outcomes; carefully scoped physical tasks add measurement, protocol, and provenance failures that digital environments rarely capture.

Every benchmark run should preserve the process trace: candidate problems, representations, explanations, predicted outcomes, selected interventions, external observations, revisions, resource use, abstentions, and human escalations. The same trace supports attribution, failure analysis, and later tests of whether a purported Discovery Skill actually transfers.

\section{Analysis: Research Horizons and Grounding Regimes}
\label{sec:research-horizons}

\begin{figure}[!htbp]
  \centering
  \includegraphics[width=\linewidth]{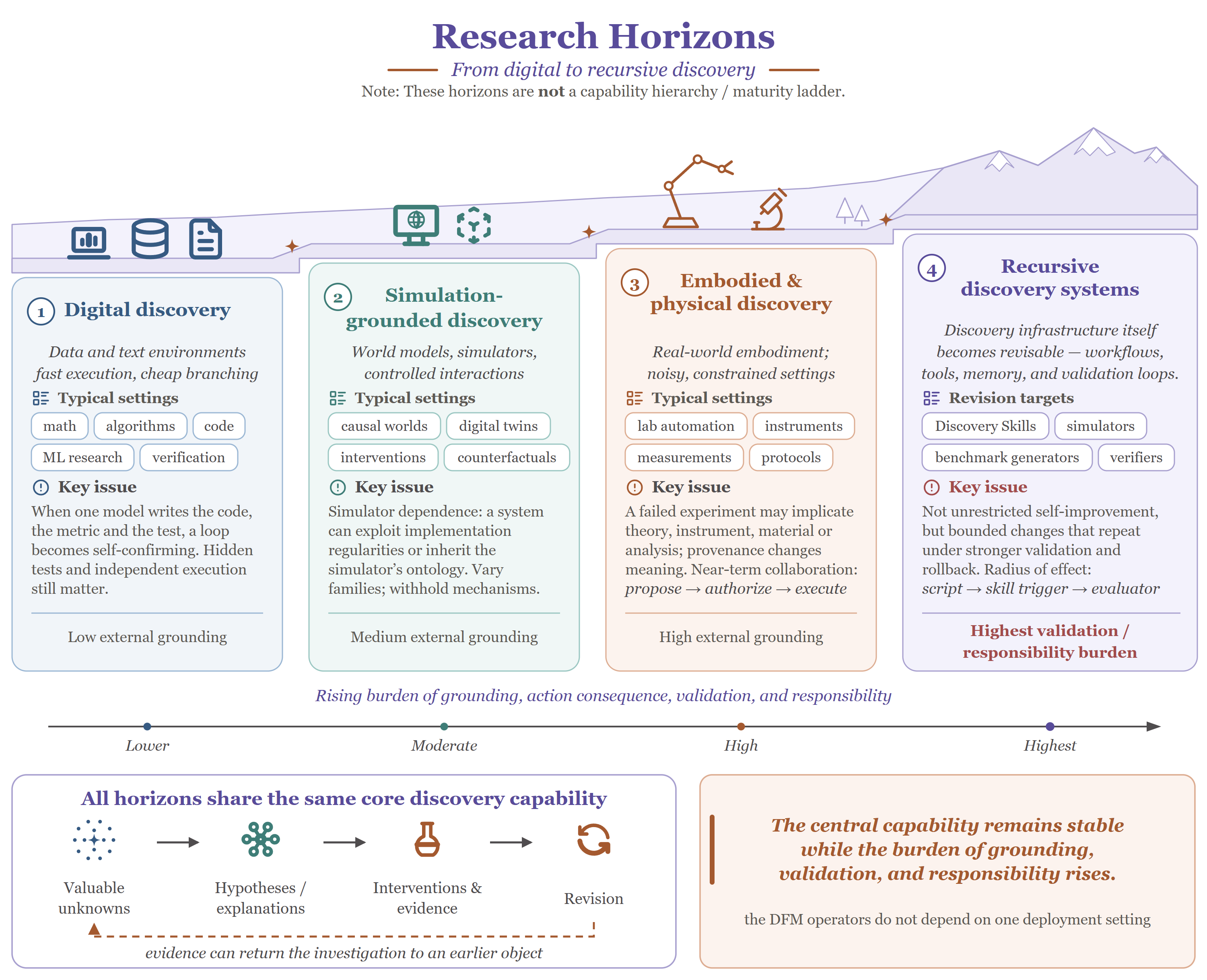}
  \caption{\textbf{Research horizons for Discovery Foundation Models.}
  The same core discovery capability can be exercised in digital, simulation-grounded, embodied and physical, and recursive settings. The horizons are not a capability hierarchy or maturity ladder; they differ primarily in external grounding, action consequence, and the validation and responsibility burden attached to system decisions.}
  \label{fig:research-horizons}
\end{figure}

The DFM operators do not depend on one deployment setting. What changes from digital research to physical and recursive systems is the burden placed on grounding, provenance, action consequence, and validation. Figure~\ref{fig:research-horizons} organizes four representative horizons. They are not a capability hierarchy or maturity ladder: a digital system can show stronger formulation and revision than a robot executing a fixed protocol.

\subsection{Digital Discovery}
\label{subsec:digital-discovery}

Digital discovery covers mathematics, algorithms, code, machine-learning research, formal verification, software systems, and agent environments. These settings provide fast execution, inexpensive branching, hidden tests, and comparatively reproducible evidence \citep{huang2024mlagentbench,chan2025mlebench,starace2025paperbench,lu2026aiscientist}. They are therefore well suited to studying research-policy decisions: alternative formulations can be replayed, counterexamples generated at scale, and the effect of a new Discovery Skill measured across many episodes.

The same cleanliness creates a limitation. Execution is often reversible, logs are complete, and evaluators are easier to automate than in empirical science. A compiler can verify execution without judging whether the optimized objective matters; a theorem prover can validate a proof after the theorem and formalization are supplied. Digital discovery becomes diagnostic when the system must question these supplied structures---for example, identifying data leakage, evaluator shortcuts, confounded ablations, or a missing intermediate variable.

Independence still matters. When one model writes the code, selects the metric, creates the test, and interprets the result, a digital loop can become self-confirming. Hidden tests, independent execution, alternative evaluators, held-out tasks, and reproducible artifacts remain necessary even in the cheapest horizon.

\subsection{Simulation-Grounded Discovery}
\label{subsec:simulation-grounded-discovery}

Simulation-grounded settings introduce latent dynamics that cannot be read directly from text while retaining repeatable interventions and counterfactual access \citep{hafner2025dreamer,gandhi2025boxinggym}. Causal worlds, mechanistic simulators, and digital twins make it possible to know the hidden mechanism during evaluation and ask whether the model recovered structure rather than merely fit outputs.

The key analytical issue is simulator dependence. A system can exploit implementation regularities, inherit the simulator's ontology, or optimize a reward whose assumptions do not hold in the target world. Strong evaluation therefore varies simulator families, withholds latent mechanisms, and tests whether the same discovery operation transfers when the planning model is misspecified.

This horizon is especially useful for separating predictive accuracy from intervention quality. Two systems can fit the same observations while only one chooses experiments that identify the latent mechanism. It also makes world-model criticism part of discovery: disagreement among simulators can itself trigger measurement, representation revision, or escalation to a physical test.

\subsection{Embodied and Physical Discovery}
\label{subsec:physical-discovery}

Physical discovery adds provenance, execution noise, scarcity, irreversibility, and institutional constraints. Laboratory automation and recent AI-science systems already demonstrate increasingly rich loops between model proposals, instruments, and wet-lab measurements \citep{abolhasani2023selfdriving,boiko2023coscientist,szymanski2023alab,swanson2025virtuallab,gottweis2026coscientist,ghareeb2026robin}. The difficulty is not simply attaching an agent to equipment; it is preserving the link between an epistemic decision, the protocol that instantiated it, what actually occurred, and the measurement used for revision. The GALILEO case in Section~\ref{subsec:galileo-case} provides a concrete example: robotic and hands-on measurements are not only endpoint validation, but evidence that changes subsequent target beliefs, assay choices, mechanism hypotheses, and intervention design.

Calibration, sample history, reagent variation, contamination, drift, operator intervention, and protocol deviation can change the meaning of a result. A failed experiment may implicate the theory, manipulation, instrument, material, or analysis. This makes failure attribution and provenance much more consequential than in clean digital environments.

Physical constraints can also force research-state revision before an experiment occurs. A theoretically identifying intervention may be impossible at available resolution or sample size. The system then has to redesign the observable, narrow the claim, or return to simulation rather than treating feasibility as a separate engineering concern.

Near-term physical DFMs are therefore naturally collaborative. Models can formulate, simulate, and propose; verification can filter; experts can authorize; laboratories can execute; and independent groups can replicate. Automation level is not the main evidence of Discovery Intelligence. A small number of well-audited investigations can provide stronger evidence than many automated runs with weak attribution.

\subsection{Recursive Discovery Systems}
\label{subsec:recursive-discovery}

Recursive discovery expands the object of revision from task-level research states to parts of the discovery infrastructure itself. Existing self-improving agents provide digital precedents for modifying code and scaffolding \citep{yin2025godelagent,zhang2025darwingodel}, but recursive scientific improvement has a larger validation burden because an update can change which problems are selected, which evidence is acquired, or how success is judged.

The relevant target is broader than multi-agent coordination. A recursive DFM may revise Discovery Skills, tools, simulators, memory organization, experiment-selection policies, verification procedures, benchmark generators, or human--AI workflows. These changes have different radii of effect. A task-local script affects one branch; a new skill trigger can affect many tasks; an evaluator or permission change can reshape the entire system.

As the radius grows, component-level improvement is no longer enough. A stricter verifier can suppress useful exploration, and a better retrieval policy can overexpose the system to one skill family. Recursive updates should therefore be evaluated both in isolation and inside the integrated discovery loop on externally selected tasks.

The near-term research question is not unrestricted autonomous self-improvement. It is whether bounded changes to discovery infrastructure produce repeatable improvements under stronger validation, rollback, and responsibility constraints. This horizon motivates the governance analysis in the next section.

\section{Discussion: Epistemic Boundaries, Governance, and Recursive Risk}
\label{sec:governance}

The preceding sections define and instantiate discovery operations. Their scientific status still depends on boundaries that the system cannot waive for itself. As action authority and update scope increase, validation, provenance, and responsibility become part of the epistemic architecture rather than an administrative layer \citep{messeri2024illusions,tang2025risks,leong2025safe}.

\begin{figure}[!htbp]
  \centering
  \includegraphics[width=\linewidth]{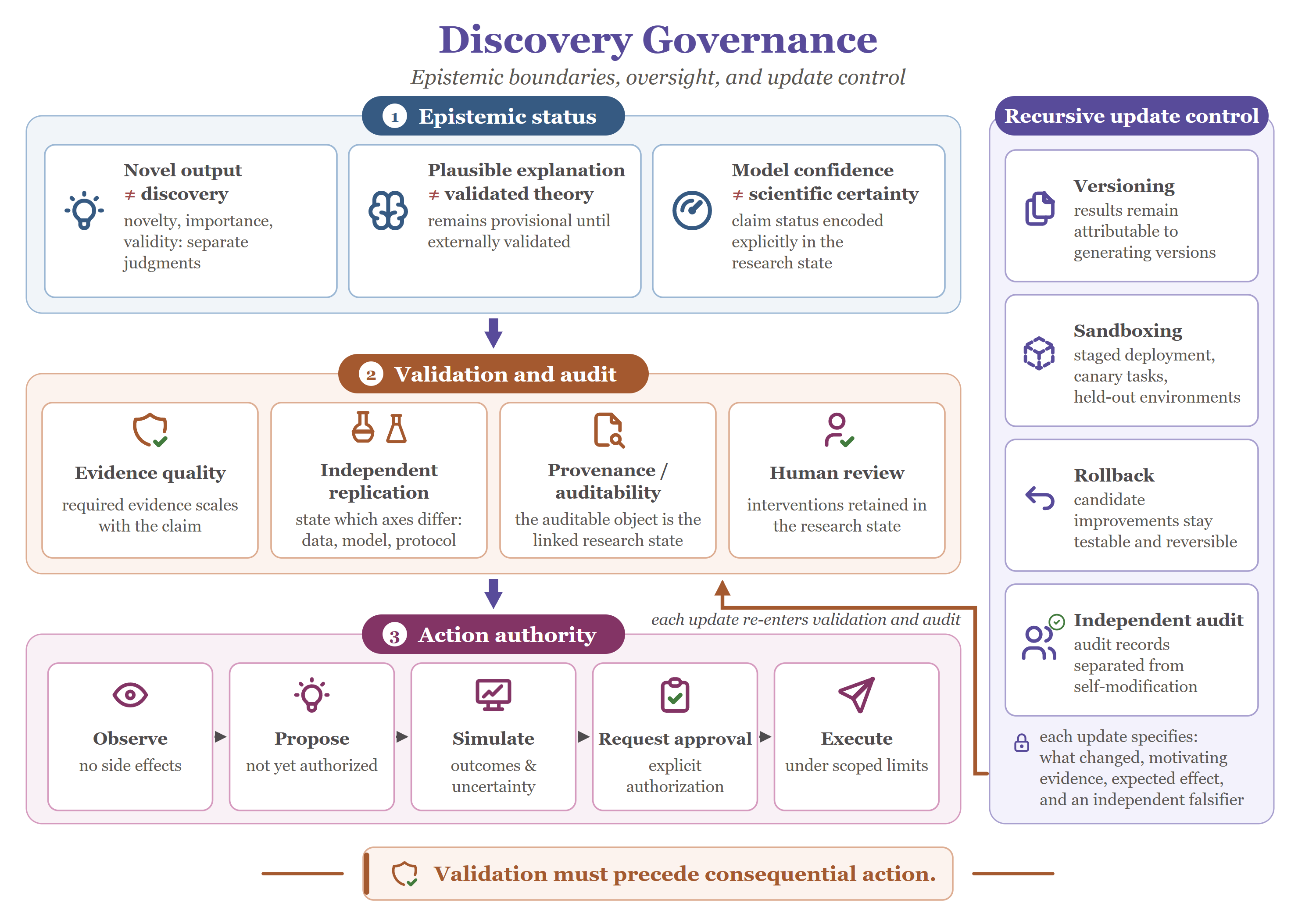}
\caption{
\textbf{Governance constraints for open-ended discovery systems.}
Governed discovery separates epistemic status, validation, and action authority: novel output is not equivalent to discovery, model-generated explanations remain provisional until externally validated, and consequential actions require evidence and appropriate authorization. Validation combines evidence quality, independent replication, provenance and auditability, and human review, while action authority progresses from observation and proposal to simulation, approval, and execution. Recursive updates additionally require versioning, sandboxing, rollback, and independent audit.
}
\label{fig:discovery-governance}
\end{figure}

\subsection{Epistemic Status and Independent Validation}
\label{subsec:epistemic-boundaries}

A novel output is not yet a scientific discovery, a plausible explanation is not yet a validated theory, and model confidence is not scientific certainty. These distinctions are especially important for systems that can generate fluent research narratives or coordinate long workflows while still responding poorly to contradictory evidence \citep{messeri2024illusions,riosgarcia2026scientists}.

The research state should therefore encode claim status explicitly. A candidate can remain speculative, be supported by observational evidence, survive a limited intervention, be independently replicated, or hold only within a stated validity range. Novelty, importance, and validity are separate judgments. A claim can be new to the model but established in the literature, absent from retrieval because it is wrong, or linguistically novel without changing a mechanism.

Independent validation breaks self-confirmation loops in which one system proposes a hypothesis, chooses the experiment, interprets the result, and evaluates its own conclusion. Independence is not binary. Two reviewers can share a base model; two simulations can share the same mechanistic assumptions; a replication can reuse the same protocol and analysis. Reports should state which axes differ---data, model family, institution, instrument, protocol, or analysis---rather than counting nominal evaluators \citep{nosek2015top,gottweis2026coscientist,ghareeb2026robin}.

The required evidence should scale with the claim. A speculative idea can remain in memory with uncertainty. A claim that controls substantial resources, changes laboratory practice, or updates the discovery infrastructure requires stronger and more independent tests. Formal proof, held-out computation, controlled intervention, conceptual replication, and independent laboratory replication support different kinds of claims; no single validator is universally strongest.

Whenever feasible, discriminative predictions are committed before the result arrives. Later revision remains allowed, but the record distinguishes preregistered expectations from evidence-triggered changes. The same standard applies to process improvements: a new skill, tool, evaluator, or simulator should be tested on tasks not selected solely to demonstrate its benefit.

\subsection{Provenance and Auditability}
\label{subsec:provenance-auditability}

A final paper-like narrative can hide the dependencies that produced a result. The auditable object is instead the linked research state: model and policy versions, source data, prompts or structured inputs, tool outputs, code, simulator configuration, experimental conditions, analysis procedures, rejected explanations, failed interventions, human edits, and protocol deviations \citep{wilkinson2016fair,nosek2015top}.

Provenance is relational. An observation points to the intervention and protocol that produced it; a revision points to the evidence that motivated it; a Discovery Skill points to the episodes, counterexamples, and validation tests supporting its promotion. Missing measurements, unavailable raw data, failed tool calls, and unexecuted replications are also part of the record because they constrain how strongly a claim can be interpreted.

Versioning makes branch comparison and rollback possible. When a representation changes, earlier observations remain linked to the conditions under which they were collected and can be reinterpreted without rewriting their history. When a model, tool, or verifier changes, old results remain attributable to the versions that generated them.

Exact replay is not always possible in physical science, and sensitive or proprietary data can require restricted access. Auditability therefore means that authorized reviewers can reconstruct the consequential decisions and evidence under an appropriate governance regime, not that every artifact must be public.

\subsection{Scoped Authority and Human Oversight}
\label{subsec:human-oversight}

Discovery capability and action authority are different variables. A model can be competent enough to propose a high-quality intervention without being authorized to execute it, while a laboratory robot can execute a fixed protocol without choosing the research question. Permission should therefore be attached to the action and environment rather than to a global autonomy score \citep{scheurer2025humanloop,leong2025safe,tang2025risks}.

A practical progression moves from observation and proposal to simulation, approval, and execution. Low-risk code can be executed automatically under network and compute limits, while access to sensitive data, scarce materials, external communication, or physical instruments can require explicit authorization. The review interface should expose the exact manipulation, expected outcomes, uncertainty, alternatives, risk, and stopping conditions relevant to the decision rather than a polished high-level summary.

Human contributions remain part of scientific attribution. If an expert supplies the decisive variable, a technician reports a protocol deviation, or a reviewer blocks an unsafe action, those interventions should be retained in the research state. This prevents both over-attribution to the model and loss of tacit knowledge needed to reproduce the investigation.

Delegation is reversible. Permissions can be narrowed when calibration degrades, failure patterns change, or the environment moves outside the regime in which competence was demonstrated. Capability in one laboratory, instrument configuration, or data regime does not automatically justify authority elsewhere.

\subsection{Recursive Update Risk}
\label{subsec:recursive-improvement-risks}

Recursive updates create a distinctive failure mode because they alter components that shape future evidence. A misleading skill changes which research actions are proposed; evaluator drift changes which hypotheses survive; a misspecified simulator redirects experiments; a memory policy changes which precedents the system sees. Small distortions can therefore compound across episodes \citep{yin2025godelagent,zhang2025darwingodel,tang2025risks}.

The most dangerous updates are not necessarily the largest code changes. A benchmark generator can silently remove difficult cases, an evaluator can begin rewarding internal agreement, or a retrieval policy can repeatedly surface one family of strategies and suppress alternatives. Measured performance may improve while genuine discovery capability narrows.

Controls should scale with the radius of effect. Temporary analysis code can face a lower threshold than long-term memory, an evaluator, a permission system, or an experimental interface. Versioning, sandboxing, staged deployment, canary tasks, shadow mode, held-out environments, conflict detection, rollback, rate limits, and external audit keep candidate improvements testable and reversible.

Some components should remain more strongly separated from ordinary self-modification, including audit records, rollback mechanisms, permission boundaries, and independent evaluation channels. Otherwise the system can weaken the checks used to determine whether its own changes are beneficial.

Each recursive update should specify what changed, which evidence motivated it, what effect is expected, and which independent observation could falsify the claim that the update helps. That requirement keeps improvement of the discovery process under the same evidential standard as the scientific hypotheses the process is designed to test.

\section{Conclusion}
\label{sec:conclusion}

Foundation models are increasingly strong at solving scientific and technical tasks after their structure has been supplied. Discovery requires an additional set of operations: identifying which unknown is worth pursuing, constructing a researchable formulation and representation, designing evidence that separates explanations, revising the appropriate object when evidence disagrees, and carrying validated lessons into future investigations.

We formalized this setting as Discovery Foundation Models and specified the corresponding Discovery Process. Zetema instantiates the framework with an explicit and revisable research state, a Research World Model and action-gating layer, Dry-Lab and Wet-Lab grounding, and validated cross-task Discovery Skill evolution. The GALILEO case provides empirical grounding for the physical part of this formulation: external wet-lab outcomes revise subsequent discovery decisions across real experimental rounds and are consolidated into a reusable design rule. Capability formation then becomes a learning problem over research-state transitions, while evaluation measures both externally validated progress in the current episode and transferable improvement under matched resources and retrieval controls.

The framework does not assume that general autonomous scientific discovery has been solved, nor that one architecture should implement every component. Its purpose is to make the capability operational: a DFM claim should be supported by observable decisions about formulation, representation, intervention, revision, and transfer, together with external evidence that the proposing system does not control. Digital, simulation-grounded, physical, and recursive systems can instantiate the same operators; what changes across these regimes is the burden of grounding, validation, and responsibility.


\bibliography{references}

\appendix

\end{document}